\documentclass[lettersize,journal]{IEEEtran}
\usepackage{amsmath,amsfonts}

\usepackage{array}
\usepackage[caption=false,font=normalsize,labelfont=sf,textfont=sf]{subfig}
\usepackage{textcomp}
\usepackage{stfloats}
\usepackage{url}
\usepackage{verbatim}
\usepackage{graphicx}
\usepackage{cite}
\usepackage{makecell}
\usepackage{amssymb}
\usepackage{multirow}
\usepackage{fancyhdr}
\usepackage{caption}  % 确保你加载了 caption 包
\usepackage[linesnumbered,ruled,vlined]{algorithm2e}

\SetCommentSty{mycommfont}
\usepackage{algorithmic}
\usepackage{pifont}
\newcommand{\cmark}{\ding{51}}%
\newcommand{\xmark}{\ding{55}}%
\usepackage{diagbox}
\usepackage{tikz}
\usepackage{ctable}
\newcommand*{\Mname}{Rec-AD}

\begin{document}

\title{Rec-AD: An Efficient Computation Framework for FDIA Detection Based on Tensor Train Decomposition and Deep Learning Recommendation Model}

\author{Yunfeng Li,~\IEEEmembership{Student Member,~IEEE}, Junhong Liu,~\IEEEmembership{Member,~IEEE}, Zhaohui Yang,~\IEEEmembership{Student Member,~IEEE} \\
Guofu Liao,~\IEEEmembership{Student Member,~IEEE}, Chuyun Zhang,~\IEEEmembership{Student Member,~IEEE}
        % <-this % stops a space

\thanks{Yunfeng Li is with the Department of Computer Science, University of California, Santa Barbara, CA, USA  (e-mail: yunfengli@ucsb.edu).}
\thanks{Junhong Liu is with the Department of Electrical and Electronic Engineering, The University of Hong Kong, Hong Kong SAR, China (e-mail: jhliu@eee.hku.hk).}
\thanks{Zhaohui Yang is with the Department of Computer Science, University of California, Santa Barbara, CA, USA (e-mail: zhaohui@ucsb.edu).}
\thanks{Guofu Liao is with the Department of Electronics and
Information Engineering, Shenzhen University,
Shenzhen, China (e-mail: liaoguofu2022@email.szu.edu.cn).}
\thanks{Chuyun Zhang is with the Department of Electronics and
Information Engineering, Shenzhen University,
Shenzhen, China (e-mail: cyzhang@szu.edu.cn).}

}



\maketitle

\begin{abstract}
Deep learning models have been widely adopted for False Data Injection Attack (FDIA) detection in smart grids due to their ability to capture unstructured and sparse features. However, the increasing system scale and data dimensionality introduce significant computational and memory burdens, particularly in large-scale industrial datasets, limiting detection efficiency.
To address these issues, this paper proposes Rec-AD, a computationally efficient framework that integrates Tensor Train decomposition with the Deep Learning Recommendation Model (DLRM). Rec-AD enhances training and inference efficiency through embedding compression, optimized data access via index reordering, and a pipeline training mechanism that reduces memory communication overhead. Fully compatible with PyTorch, Rec-AD can be integrated into existing FDIA detection systems without code modifications.
Experimental results show that Rec-AD significantly improves computational throughput and real-time detection performance, narrowing the attack window and increasing attacker cost. These advancements strengthen edge computing capabilities and scalability, providing robust technical support for smart grid security.

\end{abstract}

\begin{IEEEkeywords}
Smart Grid, FDIA, DLRM, Tensor Train.
\end{IEEEkeywords}

\section{Introduction}

\IEEEPARstart{F}{alse} 
 data injection attacks (FDIAs) represent one of the most stealthy and destructive cybersecurity threats in smart grids~\cite{mohammed2025dual}. Given the massive volume and complexity of grid data, traditional FDIA detection algorithms often incur significant computational overhead, resulting in delayed response times, excessive resource consumption, and challenges in deploying such solutions on edge devices. These limitations directly compromise the security and stability of grid operations~\cite{yu2025false}. Enhancing the computational efficiency of FDIA detection algorithms is thus essential—not only to reduce latency and resource demands, but also to strengthen the smart grid’s overall defense capabilities~\cite{fahim2025graph}.

Investigating optimization techniques that enhance FDIA detection efficiency and evaluating their impact on real-world smart grid deployments are critical. Detection systems must operate within strict temporal constraints; if detection is delayed, attackers can successfully manipulate data before being identified. For example, in SCADA systems, a detection latency exceeding the dispatch cycle (e.g., 30 seconds) can provide attackers with an exploitable time window~\cite{gasmi2024advanced}. In contrast, reducing latency to the millisecond level (e.g., 500 ms) significantly narrows this window, increasing the difficulty and cost of successful attacks~\cite{said2022new}. However, most edge devices, such as intelligent substations, are resource-constrained, while centralized methods require considerable computational power. This disparity necessitates efficient detection frameworks that support deployment on distributed nodes while minimizing reliance on centralized infrastructure.

Smart grid data is characterized by high volume, high dimensionality, heterogeneity, and multi-source origins. Traditional machine learning algorithms often struggle to extract meaningful features from such data and are prone to gradient vanishing or explosion. In contrast, deep learning models offer superior capabilities in learning complex, nonlinear,  high-order representations~\cite{jalali2021automated}. Their deeper architectures enable abstraction beyond the reach of shallow models, facilitating more accurate mapping of intricate data patterns.

Time-series signals in power systems typically contain both high- and low-frequency components. These can be decomposed into interpretable sub-sequences using methods such as Wavelet Transform~\cite{zhang2020photovoltaic}, Empirical Mode Decomposition (EMD)~\cite{behera2020comparative}, Complementary Ensemble EMD~\cite{niu2020short}, or Variational Mode Decomposition~\cite{li2021multi}, which are then processed by predictive models. However, such techniques often introduce substantial computational overhead, particularly when applied to large-scale or sparse datasets. To overcome these limitations, integrating Deep Learning Recommendation Models (DLRMs) with Tensor Train (TT) decomposition emerges as a promising direction.

DLRMs have demonstrated clear advantages in handling sparse feature learning and embedding compression~\cite{yang2020mixed, liu2021learnable}, outperforming traditional collaborative filtering methods~\cite{schafer2007collaborative, huang2021novel}. Compared to computationally intensive models like CNNs or Transformers~\cite{he2016deep, vaswani2017attention}, DLRMs primarily consist of compute-bound multilayer perceptrons (MLPs) and memory-bound embedding tables. At industrial scale, embedding tables can reach terabytes~\cite{lui2021understanding, zhao2020distributed, mudigere2021softwarehardware}, far exceeding the high-bandwidth memory (HBM) capacities (typically tens of GB) of modern GPUs~\cite{choquette2020nvidia}.

To address this bottleneck, hybrid-parallel training frameworks such as Facebook NEO~\cite{mudigere2021softwarehardware} and NVIDIA HugeCTR~\cite{hugectr} have been developed. These systems distribute embedding tables across multiple GPUs (model parallelism) while applying data parallelism to MLPs. However, the scale of industrial DLRMs often requires dozens of GPUs, resulting in high costs and energy consumption.

Two main strategies have been proposed to reduce embedding table size:  
1) Quantization, which lowers bit width but can compromise training accuracy~\cite{guan2019post};  
2) Matrix factorization, including TT-based embedding (e.g., TT-Rec~\cite{yin2021tt}), which compresses embedding tables through structured tensor decomposition. While effective in minimizing accuracy loss, existing TT-based approaches often lack optimized primitives for efficient training and inference.

Alternative methods, such as Facebook DLRM~\cite{naumov2019deep} and FAE~\cite{ebrahimzadeh2021accelerating}, employ parameter server (PS) architectures~\cite{kim2019parallax, li2014scaling}, storing embeddings in host memory and executing sparse lookups on CPUs while offloading MLP computation to GPUs. However, PS-based systems suffer from two primary bottlenecks: synchronization delays between CPU and GPU, and parameter transmission overhead~\cite{guo2021scalefreectr}, both of which require further optimization.

Despite DLRM's success in recommendation systems, its application in FDIA detection remains underexplored. This work introduces a novel integration of DLRM with TT decomposition to simultaneously improve computational efficiency and maintain detection accuracy. TT decomposition compresses high-dimensional tensors, reducing both memory and compute costs, thereby facilitating real-time model inference and training.

\begin{figure} [h] \small
    %\vspace{-30pt}
    %\setlength{\abovecaptionskip}{-0.5cm}
    %\setlength{\belowcaptionskip}{-0.5cm}
    \centering
    \includegraphics[width=0.75\linewidth]{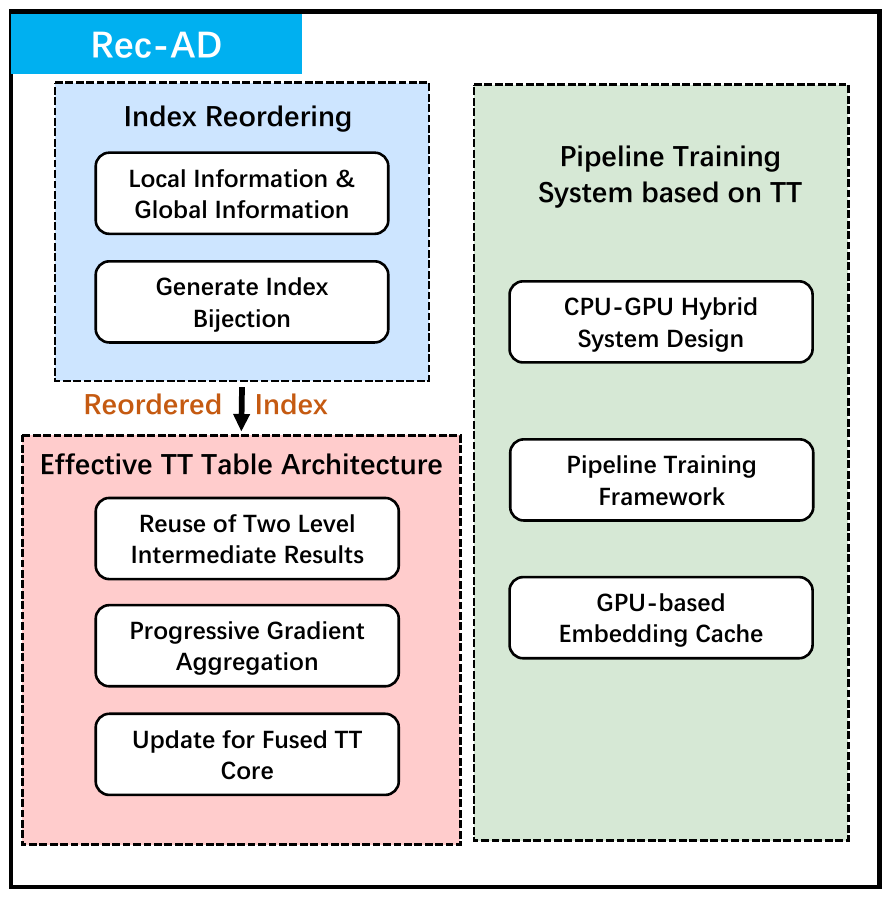}
    %\vspace{-50pt}
    \caption{Rec-AD Overview}
    \label{fig: overview}
\end{figure}

To this end, we propose the Rec-AD framework, as shown in Fig.~\ref{fig: overview}, a high-efficiency training architecture for industrial-scale DLRMs under limited GPU resources. Rec-AD adopts a multi-level co-optimization strategy across the algorithm, input, and system layers to address challenges including model complexity, data sparsity, and distributed training~\cite{ebrahimzadeh2021accelerating, hwang2020centaur, zhao2019aibox, yin2021tt}. Empirical evaluations demonstrate that Rec-AD achieves superior training speed and system efficiency, offering a deployable solution for secure and scalable smart grid operations.

Rec-AD enhances FDIA detection by improving model generalization, training/inference performance, and adaptability in distributed environments. Leveraging deep embeddings, it automatically learns high-dimensional spatiotemporal features without requiring manual feature engineering. Given the distributed and heterogeneous nature of grid data, Rec-AD is also well-suited for integration with federated learning frameworks to enable cross-region generalization.
Compared with state-of-the-art DLRM systems, Rec-AD achieves up to a 3$\times$ speedup in training time and supports large-scale deployment with limited GPU resources. These improvements enable practical, real-time FDIA detection in smart grid environments of industrial scale.

\textbf{Key Contributions:}
\begin{itemize}
  \item \textbf{Algorithm Level:} We propose the Efficient TT (Eff-TT) embedding table, which leverages TT structure for compact storage and efficient lookup. Eff-TT is fully compatible with PyTorch’s nn.EmbeddingBag() API.
  \item \textbf{Input Level:} We introduce index reordering to optimize skewed memory access patterns at both batch and table levels, improving locality and reducing cache misses.
  \item \textbf{System Level:} Rec-AD expands memory capacity using host memory and implements a three-stage pipeline to reduce CPU-GPU communication. A lightweight caching mechanism mitigates read-after-write conflicts, as shown in Table~\ref{table: framework comparision}.
\end{itemize}

\section{FDIA Detection Based on Deep Learning Recommendation Model with Tensor-Train Decomposition}
This section introduces the fundamental principles of DLRMs, TT decomposition, and embedding tables.

\begin{table}[t] \small
    % \vspace{-10pt}
    \caption{Comparison between different frameworks.}
    \centering
    \vspace{-5pt}
    \scalebox{0.78}{
    \begin{tabular}{l|c c c c}
      \toprule
        \textbf{Framework} & \textbf{\makecell[c]{Host \\Memory}} & \textbf{\makecell[c]{Embedding \\ Compression}} & \textbf{\makecell[c]{CPU-GPU \\ Comm. Latency}}  & \textbf{\makecell[c]{Compression \\ Overhead}} \\
    \hline
    \textit{DLRM}~\cite{naumov2019deep} & {\cmark} & {\xmark} & High & N/A  \\ 
    \textit{FAE}~\cite{ebrahimzadeh2021accelerating} & {\cmark} & {\xmark} & Moderate  & N/A \\
    \textit{TT-Rec}~\cite{yin2021tt} & {\xmark} & {\cmark} & N/A & High \\
    \textit{XGBoost}~\cite{9869474} & {\cmark} & {\xmark} & N/A & N/A  \\
    % \textit{HugeCTR}~\cite{hugectr} & {\xmark} & {\xmark} & N/A & N/A \\
    % \textit{TorchRec} & {\xmark} & {\xmark} & Not Involve  \\
    \hline 
    \\[-0.95em]
    \textit{\textbf{Rec-AD}} & {\cmark} & {\cmark} & Low & Low  \\
    \specialrule{.1em}{.05em}{.05em} 
    \end{tabular}}
    \label{table: framework comparision}
    \vspace{0pt}
\end{table}

\begin{figure} [h] \small
    %\vspace{-50pt}
    %\setlength{\abovecaptionskip}{-0.3cm}
    %\setlength{\belowcaptionskip}{-0.5cm}
    \centering
    \includegraphics[width=0.9\linewidth]{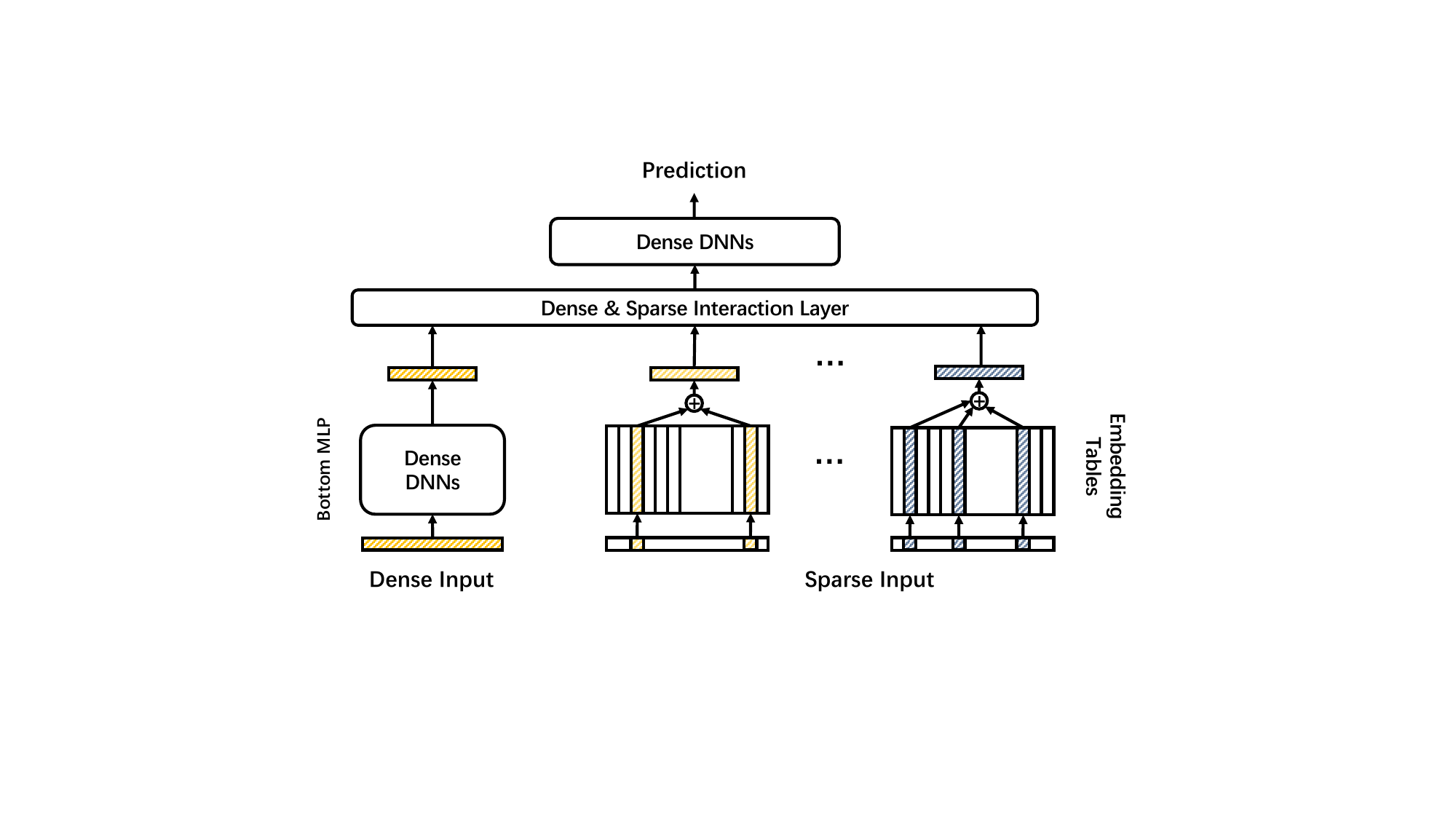}
    %\vspace{-25pt}
    \caption{DLRM framework}
    \label{fig: pre}
    \vspace{-25pt}
\end{figure}

\subsection{Deep Learning Recommendation Model}
As illustrated in Fig. \ref{fig: pre}, the Deep Learning Recommendation Model (DLRM) architecture incorporates two distinct types of input: \textbf{dense features} and \textbf{sparse features}.

The dense input is first processed by the bottom Multilayer Perceptron (MLP), which typically encodes continuous-valued features such as user age, login time, voltage magnitude, phase angle, or power readings in smart grid environments. In contrast, sparse input consists of categorical variables encoded as one-hot or multi-hot binary vectors, representing entities such as user-item interactions, bus IDs, generator identifiers, or load indices. These sparse vectors are mapped to dense, low-dimensional representations through embedding lookups that retrieve the corresponding rows from pre-trained embedding tables.

Subsequently, both dense and sparse features are projected into a shared latent space and integrated through a feature interaction layer, which computes pairwise dot products among all feature vectors to capture higher-order correlations. The output of this interaction is then fed into a top-level MLP, which performs either click-through rate (CTR) prediction or classification. In conventional recommender systems, the model outputs a CTR estimate, indicating the probability of a user clicking on a recommended item. In the context of smart grids, however, the classification output determines whether a given system state vector has been compromised by an attack.

A central challenge in DLRM training lies in the significant memory demands imposed by the embedding tables~\cite{naumov2019deep, ebrahimzadeh2021accelerating},especially in large-scale applications~\cite{lui2021understanding, mudigere2021softwarehardware}. To mitigate this issue and ensure efficient synchronization of embedding parameters, advanced memory management techniques such as software-managed embedding caches~\cite{guo2021scalefreectr} are employed. These techniques effectively reduce communication latency between GPUs' HBM and host DRAM, thereby enhancing overall training throughput.

\subsection{Tensor-Train Decomposition}\label{tt_intro}
Deep Neural Network (DNN) models can be compressed using TT decomposition, which offers high compression ratios with minimal accuracy loss~\cite{tjandra2017compressing, qu2021hardware, yin2021tt}. The TT decomposition breaks down a $d$-dimensional tensor $\mathcal{W}\in \mathbb{R}^{n_1\times n_2\times \cdots \times n_d}$ into multiplications of $d$ lower-dimensional tensors~\cite{oseledets2011tensor, wang2020adtt, novikov2020tensor}. The elements of the tensor $\mathcal{W}$ are computed using a series of summations, as shown in the following equation:

\begin{equation}\small
    \begin{split}
    \mathcal{W}(i_1,\cdots,i_k,i_d) = \sum_{r_0=1}^{R_0} \cdots \sum_{r_k=1}^{R_k} \sum_{r_{d}=1}^{R_{d}} \mathcal{D}^{(1)}(r_0,i_1,r_1) &  \\ \mathcal{D}^{(2)}(r_1,i_2,r_2)\cdots \mathcal{D}^{(d)}(r_{d-1},i_d,r_d)
    \end{split}
\end{equation}

$\mathcal{W}(i_1,\cdots,i_k,i_d)$ is a $d$-dimensional tensor, $\mathcal{D}^{(d)}(r_{d-1},i_d,r_d)$ is the $k-th$ core tensor. $r_d$ is the rank index, where $r_0$ and $r_d$ are the boundary ranks of the first and last core tensors, respectively. Where $R_k$ ($k=0,1,2...d$) represents the so-called TT ranks, which are hyperparameters, and $R_0 = R_{d} = 1$ by definition. The low-dimensional tensor $\mathcal{D}^{(k)}\in \mathbb{R}^{R_{k-1}\times n_k \times R_k}$ is referred to as the \textbf{TT core}.

It is possible to compress the embedding table~\cite{hrinchuk2020tensorized} using a generalized version of the TT decomposition. Assuming $W\in\mathbb{R}^{M\times N}$ is an embedding table, we can factorize it such that $M = m_1\times m_2 \times \cdots \times m_d$ and $N = n_1\times n_2 \times \cdots \times n_d$ for the sizes $M$ and $N$. Then, a $d$-dimensional tensor $\mathcal{W} \in \mathbb{R}^{(m_1\times n_1)\times(m_2\times n_2)\times \cdots \times(m_d\times n_d)}$ can be created using the embedding table $W$. TT decomposition can now be applied to the transformed embedding table $\mathcal{W}$. The following equation can be used to find the element in $\mathcal{W}$ indexed by $[(i_1\cdot j_1),(i_2\cdot j_2),\cdots,(i_d\cdot j_d)]$:
\begin{equation}\small\label{eq:tt}
    \begin{split}
    \mathcal{W}[(i_1\cdot j_1),(i_2\cdot j_2),\cdots,(i_d\cdot j_d)] &=   \\ \mathcal{D}^{(1)}[(i_1\cdot j_1),:]\mathcal{D}^{(2)}[:,(i_2\cdot j_2)&,:]\cdots\mathcal{D}^{(d)}[:,(i_d\cdot j_d)] 
    \end{split}
\end{equation}

\begin{figure} [h] \small
    %\begain{adjustwidth}{-\extralength}{0cm}
    %\vspace{-50pt}
    %\setlength{\abovecaptionskip}{-0.5cm}
    %\setlength{\belowcaptionskip}{-0.5cm}
    \centering
    \includegraphics[width=0.9\linewidth]{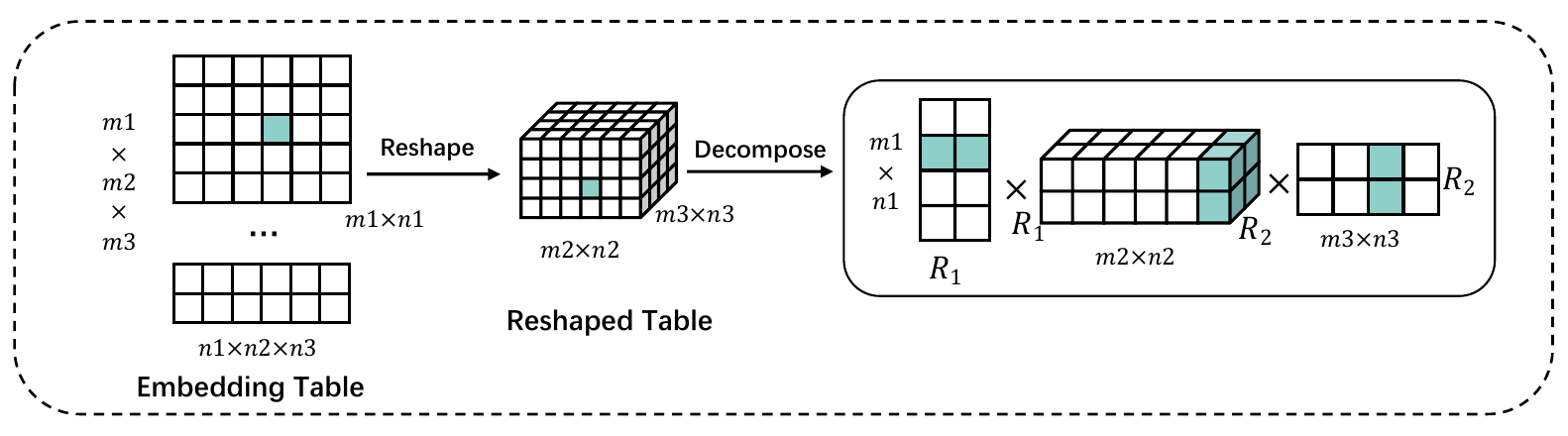}
    %\vspace{-50pt}
    \caption{The TT table is represented by an embedding table. The sizes of the decomposed embedding table are $m_i$ and $n_i$, and the TT rank is $R_i$.}
    %\end{adjustwidth}
    \label{fig: tt workflow}
    %\vspace{0pt}
\end{figure}

A decomposed embedding table is commonly referred to as a Tensor Train (TT) table. As depicted in Fig.~\ref{fig: tt workflow}, the conversion of a conventional embedding table into a TT format involves decomposing high-dimensional embeddings into a sequence of low-rank tensor cores using the Tensor Train decomposition.

While TT tables offer significant memory savings, their use in DLRM training introduces non-trivial computational challenges. Unlike standard embedding lookups supported by the PyTorch Embedding API, retrieving a single embedding vector from a TT table entails a series of tensor multiplications, which can substantially increase computational latency. This performance gap highlights the necessity of developing optimized TT-based embedding structures specifically designed for large-scale DLRM training tasks.

The trainable TT embedding table provides a compact and efficient representation of large-scale embeddings, particularly suitable for high-dimensional machine learning applications. Through chained low-rank factorization, the original high-dimensional embedding matrix is decomposed into a set of smaller tensor cores. This transformation reduces the memory complexity from exponential to linear with respect to the input dimensions, greatly enhancing scalability. As a result, TT embeddings enable efficient representation of massive embedding tables commonly used in domains such as recommender systems and natural language processing.

During model training, gradients are computed directly with respect to the TT cores, facilitating parameter updates in a memory- and compute-efficient manner. In the inference phase, the TT structure allows for fast reconstruction of embedding vectors, thereby improving model responsiveness. Overall, TT embedding tables offer multiple advantages, including reduced memory consumption, improved computational efficiency, and broad applicability to various machine learning models. These characteristics make them a powerful tool for enhancing system performance and scalability in large-scale deep learning applications.

\subsection{Data characteristics and Embedding table information}
Beyond model architecture, the characteristics of training data in DLRM exhibit distinct behaviors, particularly relevant to the optimization of TT-based embedding operators. One prominent feature is the presence of a \textbf{power-law} distribution in the activated sparse indices.

% As illustrated in Fig.~\ref{fig: global info}(a), the cumulative distribution of access frequencies across three benchmark datasets reveals that a small subset of indices accounts for the majority of embedding table lookups. Conversely, the remaining indices are accessed far less frequently, displaying a long-tail distribution. This access pattern suggests that intermediate results corresponding to high-frequency indices can be cached and reused, thereby reducing redundant computations during TT embedding lookups (see Section~\ref{sec:lookup} for further discussion).

% Moreover, the number of unique sparse indices in a batch can be disproportionately large, especially when the embedding dimensionality is lower than the batch size. Fig.~\ref{fig: global info}(b) highlights the disparity between the batch size and the average number of unique entries per batch, reflecting substantial padding overhead.

Frequent intra-batch reuse of embedding indices introduces redundant gradient computations during backpropagation. This can lead to distorted updates of the TT cores due to overrepresentation of certain gradients. This observation underscores the potential for gradient aggregation optimization, which aims to mitigate unnecessary computation during TT-based backpropagation by consolidating repeated gradient contributions. A detailed discussion of this technique is provided in Section~\ref{sec: backward}.

% \begin{figure} [t] \small
%     \centering
%     \includegraphics[width=0.9\linewidth]{figures/Re_access_pattern1.pdf}
%     \vspace{-5pt}
%     \caption{Characteristics of DLRM training data: (a) The three real-world DLRM datasets have highly skewed access patterns (embedded rows are sorted by access frequency); (b) The batch size and the average number of unique indexes in each batch vary significantly~\cite{wang2022rec}.}
%     \label{fig: global info}
%     \vspace{-5pt}
% \end{figure}

%\subsection{嵌入表信息特点}
The concept of global information refers to the access patterns observed across DLRM embedding tables, where certain rows exhibit disproportionately high popularity and frequency. These patterns frequently follow a characteristic power-law distribution, a phenomenon widely reported in various deep learning models, and are anticipated to be instrumental in future system-level optimizations~\cite{ebrahimzadeh2021accelerating, kal2021space, gupta2020deeprecsys}.

In contrast, the concept of local information, which pertains to data characteristics within subsets of training data, has received relatively limited attention. Whereas global information reflects aggregate statistical trends across the entire dataset, local information captures the relational structure of indices within specific regions or partitions. In the context of mini-batch training, local information may manifest as temporally correlated or domain-specific user behaviors, such as interactions linked to professional tasks versus leisure activities.

Leveraging both global and local information simultaneously can substantially improve the training efficiency of DLRM by enhancing data access locality and minimizing redundant computations. This dual-awareness approach enables the training pipeline to more effectively exploit the underlying structure of the input data, thereby facilitating the development of scalable and latency-sensitive deep recommendation systems.

% \newpage

\section{Efficient computing and optimization strategies}
\label{sec: eff_tt}
% In this section, we will detail our Eff-TT table design which includes forward phase intermediate result reuse and backward phase in-advance gradient aggregation.
\subsection{Design and Optimization of Eff-TT Table with Index Reordering}

In this section, we present a comprehensive overview of the design rationale underlying the Efficient Tensor Train (Eff-TT) embedding table, with a particular emphasis on the reuse of intermediate computation results during both the forward and backward propagation phases. The reuse mechanism in Eff-TT is implemented at two distinct levels, each introducing minimal memory overhead. Nevertheless, embedding lookups and gradient evaluations in the TT format involve complex tensor contractions, which can incur substantial computational costs, particularly during large-scale training. If left unaddressed, these costs may significantly prolong the overall training time.

A core innovation of the Eff-TT table lies in its ability to reuse intermediate computational results across training iterations. This reuse strategy effectively reduces memory consumption while improving computational efficiency. However, without proper optimization, the additional operations introduced by TT-based computations may become a major performance bottleneck.

To mitigate this issue, we propose index-aware reordering and sorting strategies that enhance data locality and minimize redundant tensor operations. These techniques exploit inherent structural patterns in both the input data and the tensor decomposition layout to optimize memory access and improve computation throughput. By aligning data access patterns with TT core structures, the proposed methods substantially reduce lookup latency and gradient backpropagation overhead.

The integration of intelligent indexing and sorting mechanisms leads to considerable improvements in training performance, particularly under constrained memory or real-time requirements. These optimizations enable the efficient deployment of DLRM models on resource-limited hardware, ensuring high training throughput without sacrificing model accuracy.

Overall, the proposed Eff-TT design achieves significant efficiency gains in embedding table operations, facilitating the scalable training of large recommendation models while maintaining high data utilization and minimal hardware demands.

\begin{figure} [t] \small
    \centering
    %\vspace{-50pt}
    %\setlength{\abovecaptionskip}{-0.5cm}
    %\setlength{\belowcaptionskip}{-0.5cm}
    \includegraphics[width=0.9\linewidth]{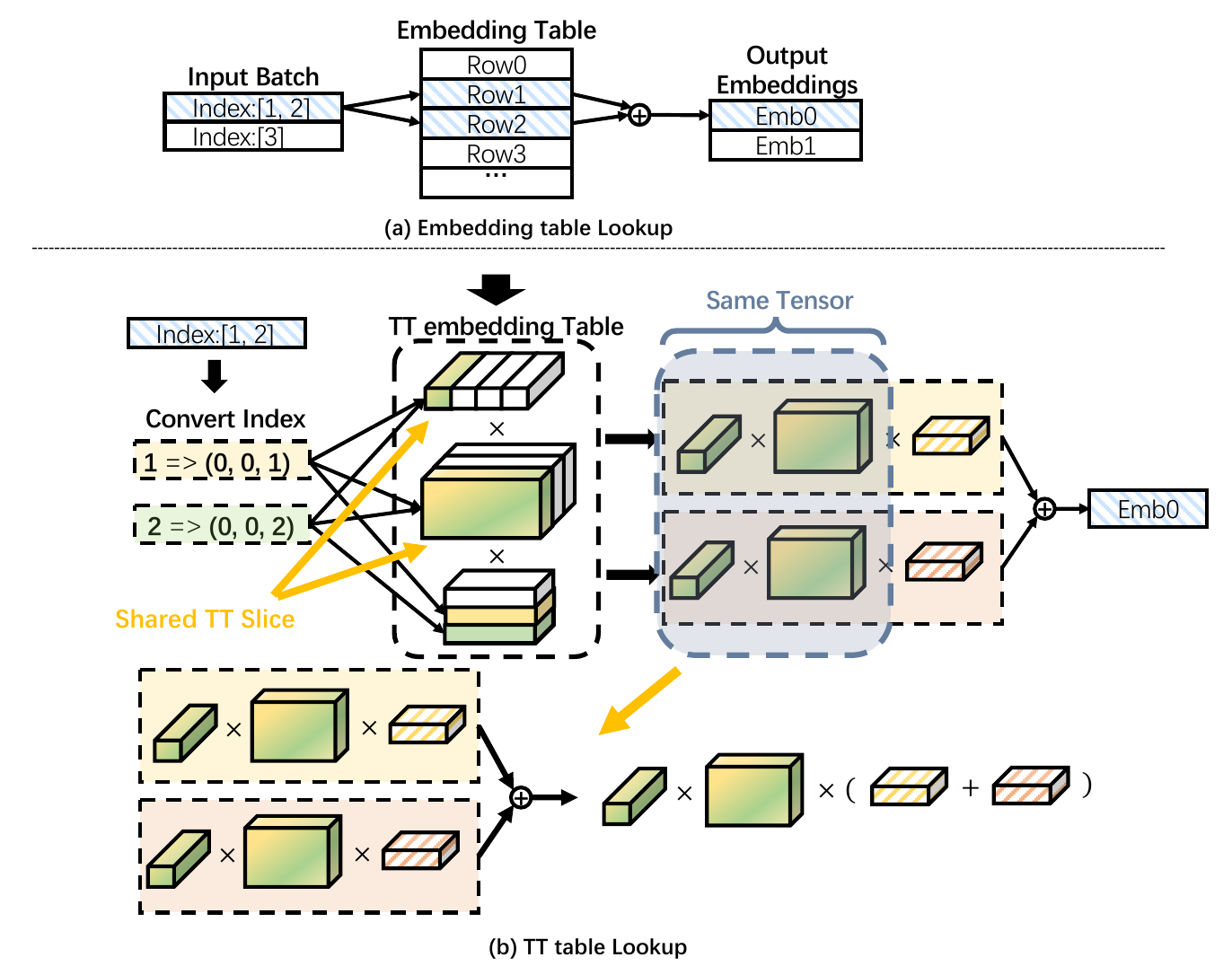}
    %\vspace{-5pt}
    \caption{The difference between TT table lookup and embedded table lookup: (a) embedded table lookup process; (b) TT table lookup process. Intermediate results can be reused through shared TT fragments and different TT indices (corresponding to different colors).}
    \label{fig: lookup}
    \vspace{-5pt}
\end{figure}

\subsection{Data Reuse}\label{sec:lookup}
Fig~\ref{fig: lookup}(a) illustrates the embedded table lookup technique. The input sample of a mini-batch contains numerous indices. Let's consider $\mathit{Index[1,2]}$ as an example. To acquire the embedding of $\mathit{Index[1,2]}$, we must first extract $\mathit{Row_1}$ and $\mathit{Row_2}$ from the embedding table to calculate $Emb_0 = Row_0 + Row_1$. Then, we add elements row-wise to $Row_1$ and $Row_2$.

This technique involves several steps for TT table lookup. Retrieving $\mathit{Index[1,2]}$ from the TT table entails a few stages. We firstly convert the indices into TT indices, as illustrated in Fig~\ref{fig:  lookup}(b). The 1-D embedding table index must first be transformed into a higher-dimensional TT index since every TT table has many TT cores, each requiring an index.

Assuming the original index is $i_{emb}$ and the embedding table size is $M\times N$ (where $M = m_1\times m_2 \times \cdots \times m_d$ and $N = n_1\times n_2 \times \cdots \times n_d$), the TT index of the $k$-th TT core can be calculated as follows:
\addtocounter{equation}{2}
\begin{equation}\small\label{eq: tt idx} \small
    \begin{split}
        i_k &= \frac{i_{emb}}{\prod_{i=k+1}^d m_i}\ mod\ m_k \\
        j_k &= {0,1,\cdots,n_{k-1}}
    \end{split}
\end{equation}
Where \textit{``mod"}  refers to the modulo operation. The original index $[1,0]$ in Fig~\ref{fig: lookup}(b) with $M=2\times 2\times 2$ is changed to $t_1 = [(0,:),(0,:),(1,:)]$ and $t_2 = [(0,:),(0,:),(0,:)]$.
% \yuke{add the goal of this step and all following steps} convert the indices $i_0 = 1$ and $i_1 = 0$ to 3-dimension TT indices $[(i_0^0,j_0^0),(i_0^1,j_0^1),(i_0^2,j_0^2)] = [(0,:),(0,:),(1,:)]$ and $[(i_1^0,j_1^0),(i_1^1,j_1^1),(i_1^2,j_1^2)] = [(0,:),(0,:),(0,:)]$. 
 Secondly, retrieving TT slices from the TT table involves accessing the appropriate slices from the TT cores based on the TT index for subsequent computation. Since $t_1$ and $t_2$ share identical values in their first two dimensions, they access the same TT slices from the first two TT cores in this case.

Next, the TT slice multiplication step is performed. This involves applying Equation~\ref{eq:tt} to each set of TT slices to reconstruct the corresponding row in the original embedding table. Let $Slice_{[p,q]}$ denote the TT slice from the $q$-th TT core indexed by $t_p$. Then, the embedding row for $t_1$ is computed as $Row_1 = Slice_{[0,0]} \times slice_{[0,1]} \times Slice_{[0,2]}$, and similarly, $Row_0 = Slice_{[1,0]} \times Slice_{[1,1]} \times Slice_{[1,2]}$ for $t_2$.

Finally, the embedding rows are aggregated by summing $Row_0$ and $Row_1$ to obtain the final embedding vector, denoted as $Emb_2$.
\begin{align}\label{eq:sample level} 
Emb_2 &= Row_0 + Row_1
    \\&=(Slice_{[0,0]} \times  Slice_{[0,1]} \times Slice_{[0,2]}) + \notag\\ &~~~~(Slice_{[1,0]} \times Slice_{[1,1]} \times Slice_{[1,2]})
\notag
\end{align}

% To reduce the computation complexity of TT table lookup, we seek the data reuse opportunity. 
Exploring the potential for data reuse to mitigate the computational complexity associated with TT table searches, consider the example of $\mathit{Index[1,2]}$ in Fig~\ref{fig: lookup}(b).

For $\mathit{Index[1,2]}$, the TT indices are $t_1 = [(0,:),(0,:),(1,:)]$ and $t_2 = [(0,:),(0,:),(0,:)]$. Since $t_1$ and $t_2$ share the same first two indices, $Slice{[0,0]} = Slice_{[1,0]}$ and $Slice_{[0,1]} = Slice_{[1,1]}$ can be substituted, as well as the intermediate result of $Slice_{[:,0]} \times Slice_{[:,1]}$. This leads to the simplification of the calculation of $Emb_2$:
\begin{align} \label{eq:sample level1} 
Emb_2 = Slice_{[0,0]} & \times  Slice_{[0,1]} \notag
\\ &\times \big(Slice_{[0,2]} + Slice_{[1,2]}\big) 
\end{align}

In Equation~\ref{eq:sample level}, the TT slice multiplication has been reduced from 4 to 2.

Consider an embedding table of dimensions $M\times N$ represented using three TT cores. Let's assume each input sample consists of $k$ indices. Additionally, let the average size of the transformed embedding table $\mathcal{T}$ in each dimension be $(m\times n)$, and let the average TT rank be denoted by $R$. This implies $M=m^3$ and $N=n^3$. In this scenario, the computational complexity associated with retrieving examples from the embedding table can be expressed as $\mathcal{O}_{emb} = \mathcal{O}((k-1)N)=\mathcal{O}(kn^3)$.

On the other hand, $\mathcal{O}{TT}=\mathcal{O}(k(2R-1)(n^2R+n^3)+(k-1)N)=\mathcal{O}(kn^2R^2)$ represents the computational complexity needed to retrieve samples from the TT table. Since $R\gg n$, $\mathcal{O}{TT}$ is substantially greater than $\mathcal{O}{emb}$. Ideally, intermediate results from each embedding row can be recycled if all TT indices in the sample show partial equality in a given dimension. The reduced complexity is $\mathcal{O}{\mathit{eff\_TT}}=\mathcal{O}(n^2R^2)$.

This finding demonstrates that the utilization of intermediate results at the sample level has a substantial impact on reducing the computational complexity associated with TT table lookups.

% which means our sample-level intermediate result reuse method can significantly reduce the embedding lookup latency of TT table.

\setlength{\textfloatsep}{10pt}% Remove \textfloatsep
\begin{algorithm}[t] \footnotesize
  \caption{TT Decomposition.}
  \label{algo: TT kernel}
  \SetAlgoLined 
  \SetKwInOut{Input}{input}
  \SetKwFor{For}{for each}{do in parallel}{end for}
  \SetKwInOut{Output}{output}
   Data preprocessing: dense feature max-min normalization.\\
  \Input{Embedding matrix:$T\in \mathbb{R}^{M\times N}$, Batched input:$\mathit{Batch\_idx}$, Reuse buffer: $\mathit{Buf}$.
%   Batched Indices: $\mathit{Batch\_idx}$, TT cores: $\mathit{TT\_cores}$ \\Reuse buffer: $\mathit{Buf}$, Buffer flag: $\mathit{Buf\_flag}$, \\Buffer tail: $\mathit{Buf\_tail}$,
%   Shape list: $\mathit{shapes}$
  }
   \Output{TT-cores:$\mathcal{D}^{(k)}\in \mathbb{R}^{R_{k-1}\times n_k \times R_k}$,Tensor-address pointers $\mathit{Pt\_a}$, $\mathit{Pt\_b}$, $\mathit{Pt\_c}$.}
    \tcc{Set up a few auxiliary variables. }
    %\ZY{These commends in algorithm can be deleted to shorten pages}
    $\mathit{Bufe\_len} = 0$, $\mathit{Bufe\_flag} = [0]$\\
    \For{$\mathit{Index}\in \mathit{Batch\_index}$}{
        \tcc{Find the row's buffer index. }
        $\mathit{Bufe\_index}$ = $\mathit{Index}\ /\ \mathit{length_3}$;
        
        \tcc{Verify the buffer's availability at $\mathit{Bufe\_index}$.}
        \If{$(\mathit{atomicCAS}(\mathit{Bufe\_flag}[\mathit{Bufe\_index}], 0, 1)==0)$}{
            \tcc{Update buffer length.}
            $cur\_\mathit{offset}$ = $atomicAdd(\mathit{Buf\_len}, 1)$;
            
            \tcc{Determine TT index0 and index1.}
            $\mathit{TT\_index1}$ = $\mathit{Bufe\_index}\ /\ \mathit{length_2}$;
            
            $\mathit{TT\_index0}$ = $\mathit{Bufe\_index}\ \%\ \mathit{length_2}$;
            
            \tcc{Update Pt\_a, Pt\_b, Pt\_c.}
            $\mathit{Pt\_a}[\mathit{Bufe\_index}]$ = $\&\mathit{TT\_cores[1]}+\mathit{TT\_index1}*\mathit{shapes}[1]$;
            
            $\mathit{Pt\_b}[\mathit{Bufe\_index}]$ = $\&\mathit{TT\_cores[0]}+\mathit{TT\_index0} * \mathit{shapes}[0]$;
            
            $\mathit{Pt\_c}[\mathit{Bufe\_index}]$ = $\&\mathit{Bufe}\ +\ cur\_\mathit{offset}\ * \ \mathit{shapes}[2]$;
        }
    }
\end{algorithm}

\textbf{Batch-data Reuse:}\label{sec: batch reuse} Batch processing of training data is a typical practice in practical applications, which increases the probability of data reuse. We extend the concept of reusing intermediate results from the sample level to the batch level.

\subsection{Parallel Kernel Design for Intermediate Reuse in Eff-TT Table}
\label{sec:tt-kernel}

One of our key findings is that when TT indices within a batch share partial consistency, the corresponding intermediate results can be reused. The primary challenge lies in identifying reusable intermediate results from the incoming batch. To address this, we design a specialized mechanism termed the Reuse Buffer, which stores intermediate products of the first two TT cores. This buffer is critical for avoiding redundant computations during forward and backward passes.

We propose a parallel pointer preparation kernel that enables the detection of unavoidable computations and prepares pointer arrays for batched GEMM operations (e.g., using cublasGemmBatchedEx). This kernel facilitates simultaneous matrix multiplications by consolidating them into a single computational routine.

The pointer lists for the batched GEMM kernel—$\mathit{Pt\_a}$, $\mathit{Pt\_b}$, and $\mathit{Pt\_c}$—are constructed via Algorithm~\ref{algo: TT kernel}. The addresses of the first and second TT cores ($\mathit{TT\_cores[0]}$ and $\mathit{TT\_cores[1]}$) are stored in $\mathit{Pt\_a}$ and $\mathit{Pt\_b}$, respectively, while $\mathit{Pt\_c}$ stores the result of their multiplication. The number of threads equals the number of indices in the embedding stack, with each thread processing a single embedding index (Line 2).

The length of the final TT core ($\mathit{length\_3}$) is computed (Line 3), and a reuse index is derived by dividing the index by a constant according to Equation~(3-3), which serves as the lookup address in the \textit{Reuse Buffer}. Each thread then checks whether it can skip the redundant computation (Line 4). This decision is recorded in the boolean array $\mathit{Bufe\_flag}$. If $\mathit{Bufe\_flag[Bufe\_index]}==1$, it indicates that the intermediate result has already been computed by another thread and stored in the buffer. If not, the addresses for $\mathit{Pt\_a}$, $\mathit{Pt\_b}$, and $\mathit{Pt\_c}$ must be allocated (Lines 8–10) for further GEMM computation.

Once all pointers are prepared, the batched GEMM kernel is invoked, using the arrays $\mathit{Pt\_a}$, $\mathit{Pt\_b}$, and $\mathit{Pt\_c}$ to compute and store intermediate results for the first two TT cores in the \textit{Reuse Buffer}. 

By helping the Eff-TT table selectively determine which computations are necessary, Algorithm~\ref{algo: TT kernel} significantly reduces the lookup cost in TT-based embedding tables.

\subsection{Optimization in Backward}\label{sec: backward}
During the TT table backward process, the primary tasks involve updating the TT cores' parameters and calculating their gradients. Consider a TT table with $d$ TT cores, where the following formula computes an embedding row: $e = \mathcal{D}^{(1)}[i_1,:] \mathcal{D}^{(2)}[:,i_2,:] \cdots \mathcal{D}^{(d)}[:,i_{d}] $. Initially, we extract the gradient of embedding $\frac{\partial L}{\partial e}$ during the backward process. Subsequently, the gradient of the $k$-th TT cores $\frac{\partial L}{\partial \mathcal{D}^{(k)}[:,i_k,:]}$ is calculated using the chain rule:

\begin{align}\small \label{eq: backward} 
    \frac{\partial L}{\partial \mathcal{D}^{(k)}[:,i_k,:]}  = \Big( \mathcal{D}^{(1)}[i_1,:]   \mathcal{D}^{(2)}[:,i_2,:] \cdots \mathcal{D}^{(k-1)}[:,i_{k-1}&,:] \Big)^T \cdot \notag\\  \frac{\partial L}{\partial e} \cdot \Big( \mathcal{D}^{(k+1)}[:,i_{k+1},:]  \cdots \mathcal{D}^{(d)}[:,i_{d}] \Big)^T
\end{align}

\begin{figure} [t] \small
    \centering
    %\vspace{-75pt}
    %\setlength{\abovecaptionskip}{-1cm}
    %\setlength{\belowcaptionskip}{-1cm}
    \includegraphics[width=1\linewidth]{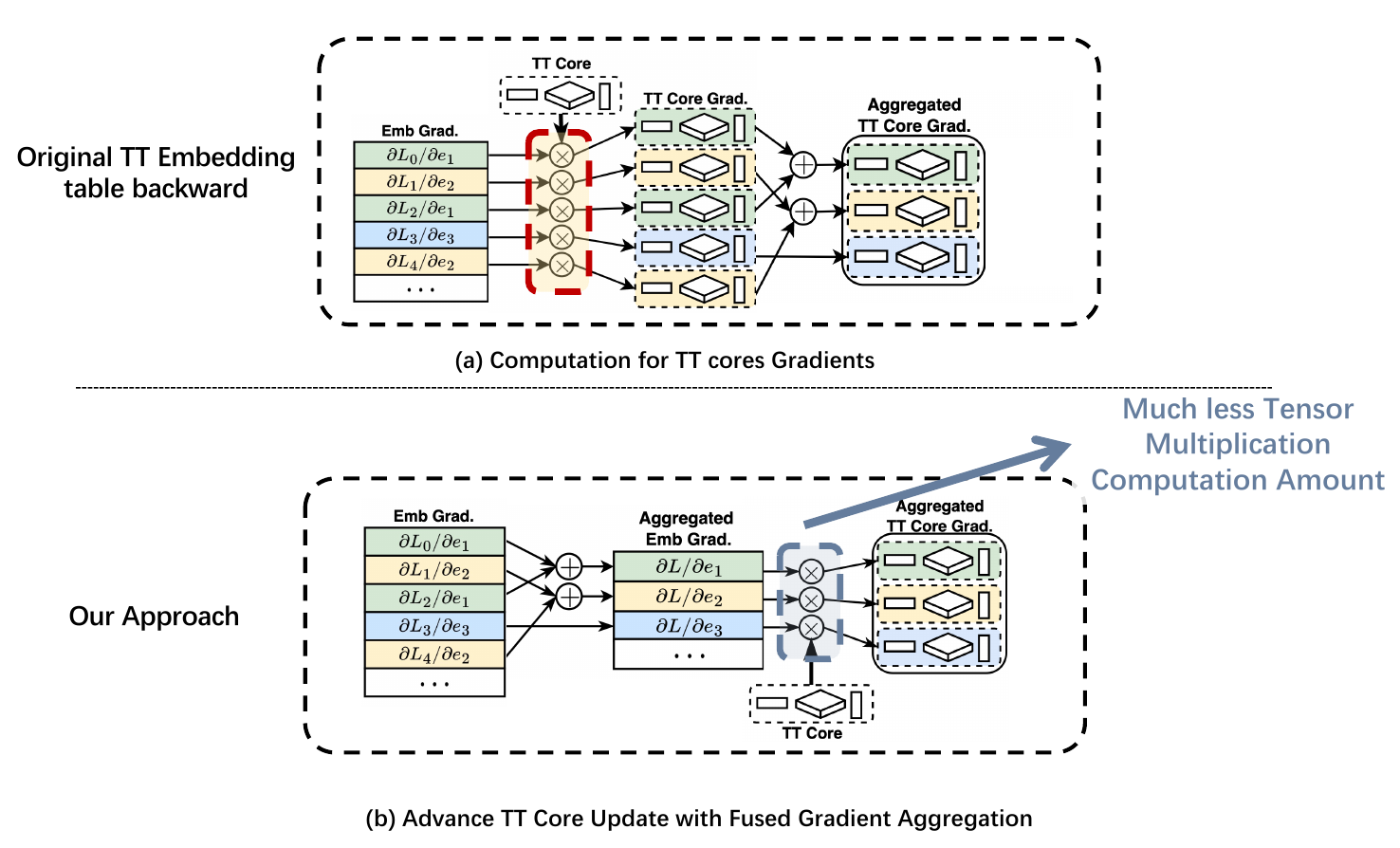}
    %\vspace{-5pt}
    \caption{TT table back propagation includes two steps: (a) TT-Rec TT table back propagation; (b) Eff-TT table back propagation and advance gradient aggregation. The advance gradient aggregation method greatly reduces tensor multiplications, and the gradients corresponding to different embeddings are represented by different colors.}
    \label{fig: unique}
    % \boyuan{Add 1-sentence high-level summary on why there are benefits.}
    \vspace{5pt} 
\end{figure}

Equation~\ref{eq: backward} illustrates that $(d-1)$ tensor multiplications are required when computing the gradient of the TT kernel, each multiplied $(d-1)$ times. With a total of $d$ TT cores, the computational complexity can be expressed as $d$ multiplied by the TT table lookup. The procedure for calculating the TT core gradient is depicted in Fig~\ref{fig: unique}(a). The gradient of the TT kernel associated with each embedding row is derived by multiplying the embedding gradient $\frac{\partial L}{\partial e}$. Next, the computed gradients are aggregated into the corresponding TT kernels. Finally, the aggregated gradients of the tensor train (TT) cores are sent back and utilized to update the TT cores. The significant number of tensor multiplications presents the primary constraint in the backward process of the TT table. To enhance the efficiency of the TT table in the backward direction and reduce redundant computations, we propose two methods as follow.
% \textit{Unique Gradient Computation} method and \textit{Fused TT-core update}. 
% illustrated in Figure~\ref{fig: unique}(b):

%  \begin{figure*} [t] \small
%     \centering
%     %\vspace{-75pt}
%     \setlength{\abovecaptionskip}{-0.3cm}
%     %\setlength{\belowcaptionskip}{-0.5cm}
%   {\includegraphics{figures/new_new_reordering11.pdf}
%     \vspace{10pt}
%     \caption{(a) Eff-TT table query without reordering; (b) Eff-TT table query with index reordering.
% For Eff-TT table, index reordering increases the chance of reusing intermediate results and improves the locality of data.
%     % Data locality improved with index reordering which provides more 
%     }
%     % \boyuan{I can understand index reordering might be helpful. However, there are two issues in this figure: 1) before and after reorder, same digits are used (0-7), which is confusing; 2) may given more intuition on why there are benefits.}
%     \label{fig: reorder}
%     \vspace{-5pt}}
% \end{figure*}

\begin{figure}[t]   % 单栏时用 figure；双栏可用 figure*
  \centering
  % 等比缩小到 0.8×
  \scalebox{0.4}{\includegraphics{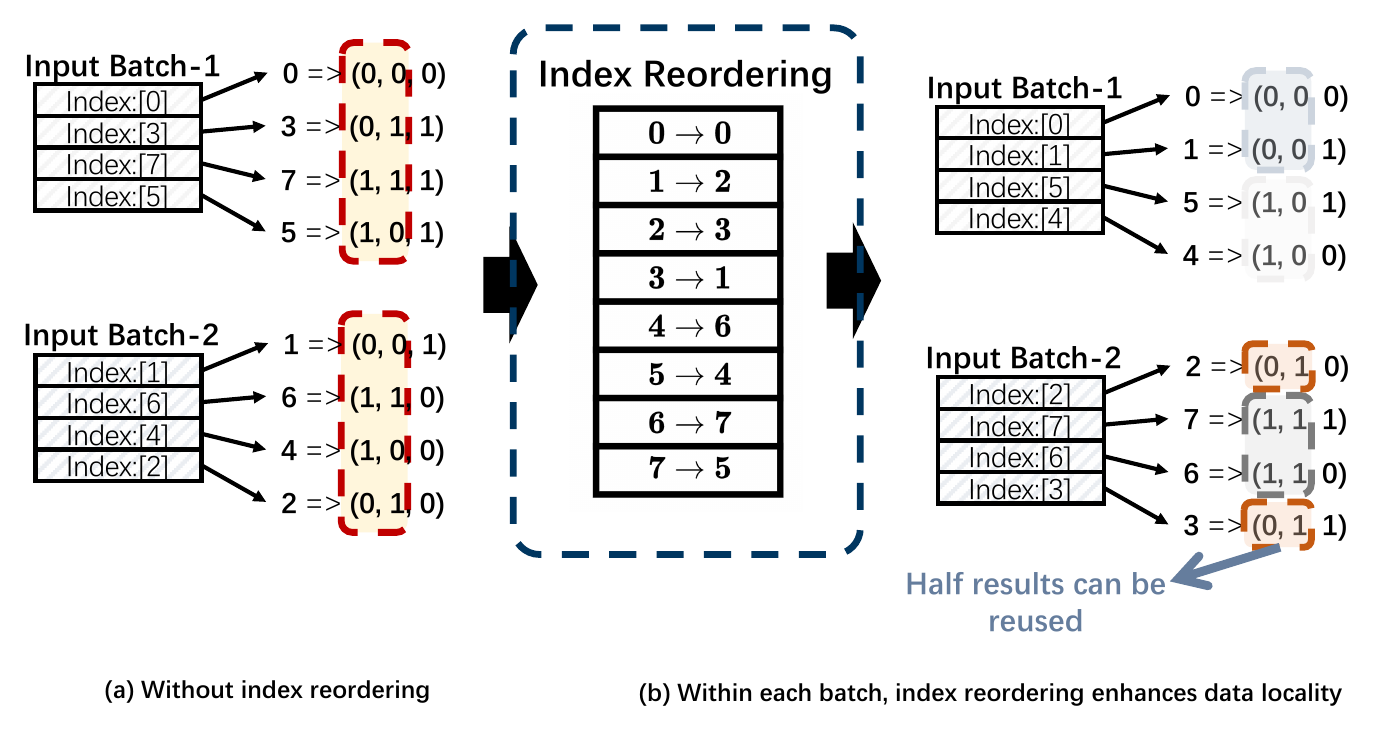}}
  \caption{(a) Eff-TT table query without reordering;
           (b) Eff-TT table query with index reordering.
           Index reordering increases the chance of reusing intermediate
           results and improves data locality.}
  \label{fig: reorder}
\end{figure}

\subsection{Gradient Aggregation Optimization}
\label{sec: unique}

 We observe that the access pattern of embedding tables is highly skewed—certain frequently used embeddings are likely to be accessed only once per batch. This skewness is reflected by the repeated appearance of similar gradient patterns across embeddings, as shown by matching gradient colors in Fig.~\ref{fig: unique}(b).

Additionally, certain embeddings may appear multiple times within a single mini-batch. To handle this redundancy, we first compute the gradients of all distinct embeddings in the index stack. Then, the gradients corresponding to the same embedding row are merged together. In Fig.~\ref{fig: unique}(b), this initial embedding gradient computation is denoted as first step. The following step  involves multiplying the aggregated embedding gradients with the TT cores to obtain the aggregated TT core gradients.

This gradient aggregation strategy significantly reduces computational overhead and also minimizes memory usage associated with intermediate tensor multiplications. By eliminating the need to compute and store redundant unaggregated data, the approach improves the overall efficiency of TT core gradient computation.

\subsection{Fused TT Core Update Optimization}
\label{sec: fused}

This optimization focuses on the final stage of TT table backpropagation, where the computed TT core gradients are returned to the GPU and stored in global memory for use by the optimizer.

In previous work such as TT-Rec~\cite{yin2021tt}, \textit{kernel fusion} is adopted to improve performance. However, it requires first aggregating TT core gradients and then performing additional data copies for parameter updates.

In contrast, our method performs early-stage gradient aggregation to directly compute the aggregated TT gradients during the backward pass. This enables fused TT core updates without requiring extra data movement. As shown in  Fig.~\ref{fig: unique}(b), the update is performed by incorporating historical information into a single fused operation.

This approach is designed to reduce the volume of data retrieved from GPU global memory and lower the launch overhead associated with CUDA kernels. As a result, it improves both computational efficiency and memory throughput in TT-based embedding training.

% 为了更清楚地展示图~\ref{fig: unique} 的推导过程，我们添加了一些解释。

% \textbf{(a) TT 表反向传播}
% 计算不同嵌入模型的梯度 \(\partial L_i / \partial e_j\). 这一步包括通过前向传播和后向传播计算损失函数 \(L\) 相对于嵌入参数 \(e_j\) 的偏导数。
% 将嵌入梯度乘以 TT 核心，计算出 TT 核心梯度。这一步骤包括计算每个 TT 核心的梯度：
% $
%  {TT Core Gradients} = {TT Core} \times (\partial L_i / \partial e_j)   
% $.
% 汇总所有 TT 核心梯度，得出总体 TT 核心梯度：
% ${TT Core Gradients Aggregated} = \sum {TT Core Gradients}$.

% \textbf{(b) 采用预先梯度聚合的 Eff-TT 表后向传播}
% 对不同嵌入的梯度进行预聚合，以减少冗余计算。这一步骤大大减少了后续计算中的张量乘法次数：
% $
%   {Aggregated Embedding Gradients} = \sum (\partial L_i / \partial e_j)  
% $.
% 使用聚合嵌入梯度计算 TT 核心梯度，减少计算负荷：
% $
%     {TT Core Gradients} = {TT Core} \times {Aggregated 
%   Embedding  Gradients}
% $.
% 使用汇总的 TT 核心梯度更新 TT 核心：
% $
%  {Updated TT Core} = {TT Core} - \eta \times {TT Core Gradients Aggregated}   
% $. $\eta$ is a hyperparameter.
% 这种方法通过预聚合梯度，有效减少了张量乘法的次数，从而提高了大规模数据的训练效率。

% \subsection{根据位置信息重建索引}
% 在本节中，我们将首先研究全局和局部的概念。然后介绍索引和排序方法，以提高 Eff-TT 表的性能。

\begin{figure} [t] \small
    \centering
    %\vspace{-75pt}
    %\setlength{\abovecaptionskip}{-3cm}
    %\setlength{\belowcaptionskip}{-3cm}
    \includegraphics[width=1\linewidth]{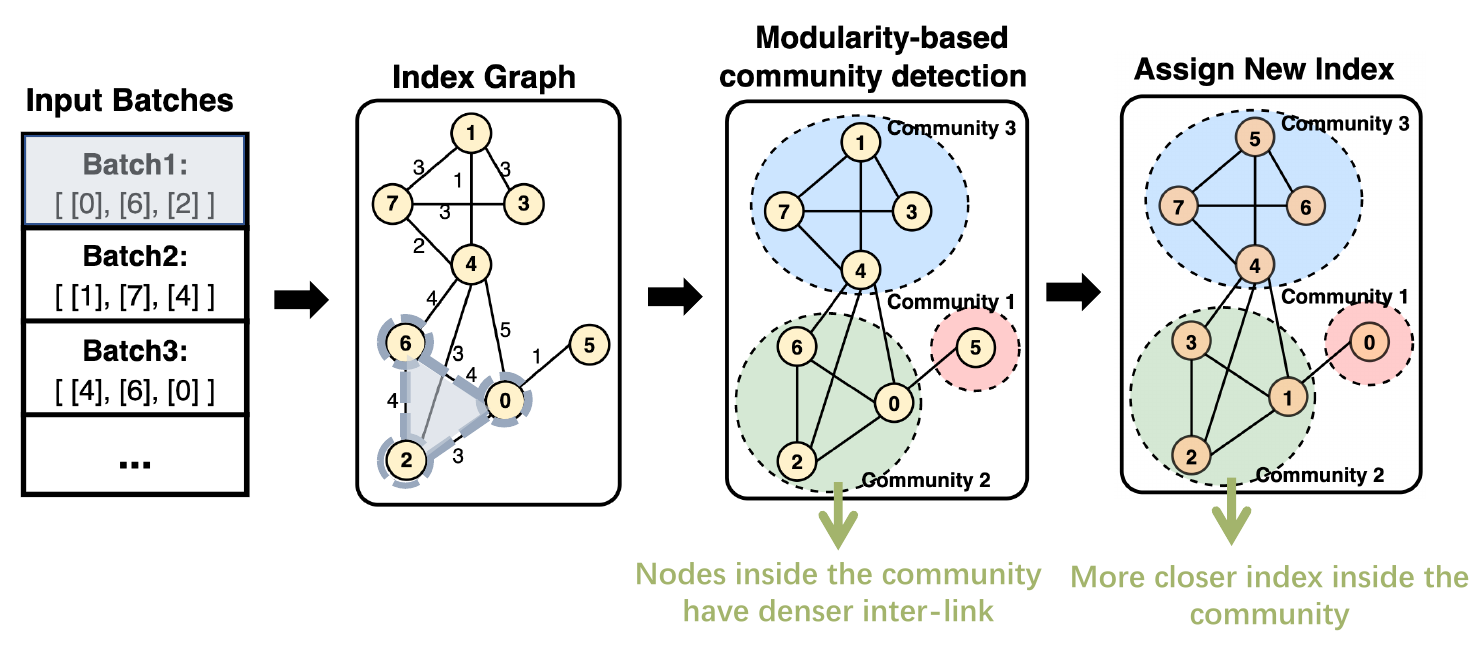}
    \vspace{-5pt}
    \caption{ The process of creating an index bijection is: 1) Create an index graph using batched input data; 2) Find communities in the graph and create a new index.}
    \label{fig: graph}
    % \vspace{-5pt}
\end{figure}

% \subsection{Locality-based Index Reordering}
% The performance of the Eff-TT table depends on the data distribution within each data batch. Based on Equation~\ref{eq: tt idx}, if the indices within a batch are closer, then more TT indices will be partially equal which will provides more opportunities for intermediate result reuse. And higher data locality is also beneficial for improving the GPU L1/L2 cache hit rate.
% And higher data locality means more overlap within the data loading from global memory. It is beneficial for improving the GPU L1/L2 cache hit rate.
% The more indices have the same $Buf\_idx$, the more intermediate result can be used, thus making TT table run faster. It demands better locality of the data in each mini batch, in other word, it would be better if the rows that are indexed in the same batch lay closer.

\begin{algorithm}[t] \footnotesize
  \caption{Index Graph Generator for training.}
  \label{algo: graph}
\SetAlgoLined
  \SetKwInOut{Input}{input}
  \SetKwInOut{Output}{output}
  \Input{Indices in Batch: $\mathit{Bat\_list}[\ ]$, \\Ordering based on frequency: $\mathit{Freq\_order}[\ ]$}
  \Output{List of Index Graph Edges: $\mathit{Edge\_list}[\ ]$}
    \tcc{Determine the threshold for hot embedding.}
    $\mathit{Hot\_threh}$ = $\mathit{Table\_len}\ *\ \mathit{Hot\_ratio}$;
    
    \tcc{Repeat with each batch.}
    \For{$\mathbf{each}$ $\mathit{Batch}$ $\mathbf{in}$ $\mathit{Batch\_list}$}{
        \tcc{Global Information: obtain an index based on frequency.}
        $\mathit{Freq\_batch}$ = $\mathit{Freq\_order}$ [$\mathit{Batch}$];
        
        \tcc{Save the index of Hot embeddings.}
        $\mathit{Freq\_batch}.\mathit{clamp}(min=\mathit{Hot\_threh})$ - $\mathit{Hot\_threh}$;
        
        \tcc{Generate edges for $\mathit{Batch}$ in the local information.}
        $\mathit{Batch\_edges}$ = $\mathit{Freq\_batch}.\mathit{self\_combinations()}$;
        
        \tcc{Append to $\mathit{Edge\_list}[\ ]$.}
        $\mathit{Edge\_list}.\mathit{append}(\mathit{Batch\_edges})$;
    }
    The output layer uses Sigmoid when the training dataset is a recommender system dataset and MSE loss function when the dataset are PV datasets.
\end{algorithm}

% \vspace*{5pt}
%\subsection{Index sorting and index bijection mechanism}
\subsection{Index Reordering for Enhanced Data Locality in Eff-TT Table}

The performance of the Eff-TT table is highly dependent on the distribution of indices within each data stack. As shown in Equation~\ref{eq: tt idx}, in the context of TT decomposition, higher adjacency among indices within a batch implies a greater number of similar (and thus reusable) TT index values.

Due to overlapping data access patterns or shared computational tasks, intermediate results can often be reused, increasing data reuse rate. This reuse not only enhances the spatial locality of GPU memory access but also improves L1/L2 cache hit rates. Without index sorting, the TT-based table lookup process proceeds as shown in Fig.~\ref{fig: reorder}(a). In this default case, no assumptions are made about the order of indices in a batch, and reusable computation opportunities are not explicitly exploited.

In practice, we observe that poor data locality in TT-based embedding lookups can have a negative impact on training performance. To address this issue, we propose physically reordering each batch’s data layout to improve within-batch access locality. While row-wise reordering of the embedding table itself is theoretically possible, the size of these tables makes such relocation costly in terms of memory movement and space overhead.

Instead, we adopt a more practical and efficient strategy: index reordering. Embedding tables are typically initialized randomly at the beginning of training, meaning all embedding rows are initially equivalent. Based on this observation, we propose a position-based index growth sorting method that enhances the spatial locality of data access within each stack.

As shown in Fig.~\ref{fig: reorder}(b), sorted indices lead to improved reuse of intermediate results. Each batch is represented as a set of indices, defined as $\mathit{Batch}_i = \{i_1, i_2, \cdots, i_n\}$, where all $i_j$ are mapped to unique identifiers via a bijective function. We denote the reordered index set as:
\begin{equation}
\mathit{New\_Batch}_i = \{\tilde{i}_j \mid \tilde{i}_j = f_{\mathit{index}}(i_j),\ i_j \in \mathit{Batch}_i\}
\end{equation}
This transformation reorganizes indices prior to embedding lookup. The result is a significant improvement in data locality within each batch, thereby increasing the likelihood of intermediate result reuse and improving computational efficiency of the Eff-TT table.

%\subsection{索引双射}\label{bijection}
\subsection{Index Bijection via Local-Global Information Fusion}
\label{sec: index-bijection}

To enhance the efficiency of index sorting and improve data locality, we propose the construction of an index bijection register. This bijection leverages both local and global statistical properties of the training data and is referred to as a dual-projection indexing strategy. The overall process is illustrated in Fig.~\ref{fig: graph}, where the bijection mechanism consists of two core components.

To extract local information from the training data, we first convert the stacked indices into an index graph. The first step is to sort the indices based on their frequency of access, as described in Algorithm~\ref{algo: graph}. The index list is ordered in descending order of access frequency, and a threshold parameter $\mathit{Hot\_ratio}$ is used to determine the set of frequently accessed embeddings (i.e., ``hot embeddings''). These embeddings are exempt from further reordering.

Subsequently, global information is used to group dynamic or cold embeddings, and the remaining prominent indices are used to construct the index graph. For each index, edges are added between two nodes if the corresponding indices co-occur in the same mini-batch. This co-occurrence relationship forms the basis for edge construction in the graph, which follows the procedure outlined in Algorithm~\ref{algo: graph}.

As shown in Fig.~\ref{fig: graph}, a positional representation is created for each index using the batch-wise context information. This is followed by arranging the indices based on a clustering criterion derived from graph connectivity. Specifically, we apply a modularity-based community detection algorithm~\cite{arai2016rabbit, zhou2019novel} to identify well-connected subgraphs (communities). The modularity metric~\cite{shiokawa2013fast} is defined as:

\begin{equation}
Q = \frac{1}{2m} \sum_{i}\left[e_{ij} - \frac{k_i^2}{2m}\right]
\end{equation}

where $e_{ij}$ denotes the total number of edges between communities $i$ and $j$, $k_i$ is the sum of degrees of nodes in community $i$, and $m$ is the total number of edges in the graph. A higher modularity score $Q$ implies denser connections within communities and sparser connections between them.

Once the graph is partitioned into communities, a new index assignment is generated. As illustrated in Fig.~\ref{fig: graph}, the indices belonging to the same community are grouped together and assigned sequentially. This dual-projection bijection utilizes both global (frequency-based) and local (batch co-occurrence) signals to maximize data reuse and cache locality.

Importantly, several steps in this indexing optimization pipeline—such as hot index identification and community detection—can be performed offline prior to training. This significantly reduces runtime overhead and ensures efficient index mapping without interrupting the training process.

% \setlength{\textfloatsep}{1pt}% Remove \textfloatsep

% \section{TT-based Pipeline Training System}\label{sec: pipeline}
% In this section, we will first detail our TT-based pipeline training system design, then we will introduce how to solve the read-after-write conflict in pipeline DLRM training with a GPU side embedding cache design.

\section{Pipeline training mechanism}\label{sec: pipeline}
This section provides a preliminary overview of the TT-based pipeline training framework, outlines the design of the pipeline training system and its solution to resolve read-after-write conflicts in DLRM pipeline training. In addition, this section discusses the design of the GPU-side embedded cache.

\begin{figure} [htb] \small
    \centering
    %\vspace{-75pt}
    %\setlength{\abovecaptionskip}{-0.5cm}
    %\setlength{\belowcaptionskip}{-0.5cm}
    \includegraphics[width=1\linewidth]{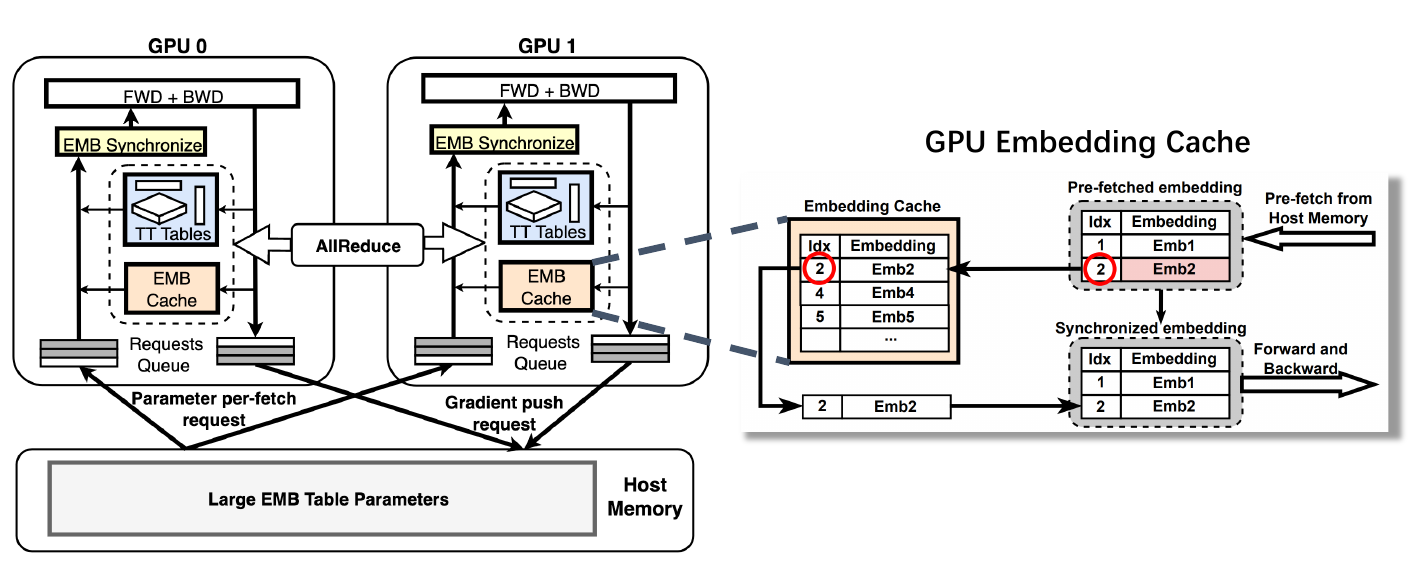}
    \vspace{-3pt}
    \caption{Pipeline training system based on TT}
    \label{fig: dataflow}
    % \vspace{-5pt}
\end{figure}

\subsection{Pipeline training system framework}
% \subsection{Hierarchical Memory Design and Pipeline Training for Scalable Embedding}
\label{sec: pipeline-training}

Building upon TT decomposition, we further propose a hierarchical memory system to enhance the scalability of Rec-AD by leveraging host memory expansion. This strategy is particularly critical when the TT-based embedding table remains too large to fit into the available GPU High Bandwidth Memory (HBM). During industrial-scale DLRM training, when data must be fed into GPU memory in real time, certain embedding table parameters are offloaded to host memory as a necessity.

However, this layered memory design imposes significant communication pressure on the GPU-CPU interface, often resulting in degraded training performance. To address this issue, we introduce pipeline training based on TT decomposition, a mechanism designed to mitigate communication latency. The overall system architecture is illustrated in Fig.~\ref{fig: dataflow}, which follows a Parameter Server (PS) paradigm suitable for scalable embedding training.

Fig.~\ref{fig: dataflow} presents the TT-based pipelined training workflow. This system manages large-scale embedding table parameters via a series of steps: synchronizing TT tables from host memory, caching them for forward and backward propagation, computing gradients, and synchronizing gradients across devices via ALLReduce before updating the TT cores. This design minimizes redundant computation, enforces gradient consistency across distributed training units, and ultimately improves overall training performance.

The top and bottom MLP layers are replicated and trained in data-parallel fashion across all GPUs. The embedding layer is split into two parts: a TT-based table that can represent the majority of embedding parameters, which is shared across GPUs. The training procedure for the FDIA detection task is summarized in Algorithm~\ref{algo: train}. The algorithm processes dense normalized features and sparse one-hot encoded features, together with target labels (e.g., attacked or clean data), to effectively learn a robust classifier.

In scenarios where GPU HBM cannot accommodate all embedding parameters, the excess parameters are retained in host memory. In this case, the CPU assumes the role of a parameter server, responsible for injecting host memory values into the prefetch queues of multiple GPUs and scheduling the next batch’s embedding parameter fetch.

GPUs serve as computing workers, processing TT-based embeddings in a round-trip fashion. These workers first gather embedding data from TT tables and prefetch queues to construct input representations. Embedding batches are then processed with support from the embedding cache and chained embedding synchronizers to ensure security and consistency across components (see Section~\ref{sec: cache design} for details).

The execution steps are as follows:
\begin{itemize}
    \item  Embeddings are updated using data-parallel training to reduce TT table gradients via forward and backward propagation.
    \item  The cache is populated before modifying the TT parameters to ensure consistency.
    \item  The server retrieves gradients and updates the host memory embedding parameters, then transfers the gradients into a historical queue for reuse or analysis.
\end{itemize}

\begin{figure} [t] \small
    \centering
    %\vspace{-75pt}
    %\setlength{\abovecaptionskip}{-0.5cm}
    %\setlength{\belowcaptionskip}{-0.5cm}
    \includegraphics[width=0.9\linewidth]{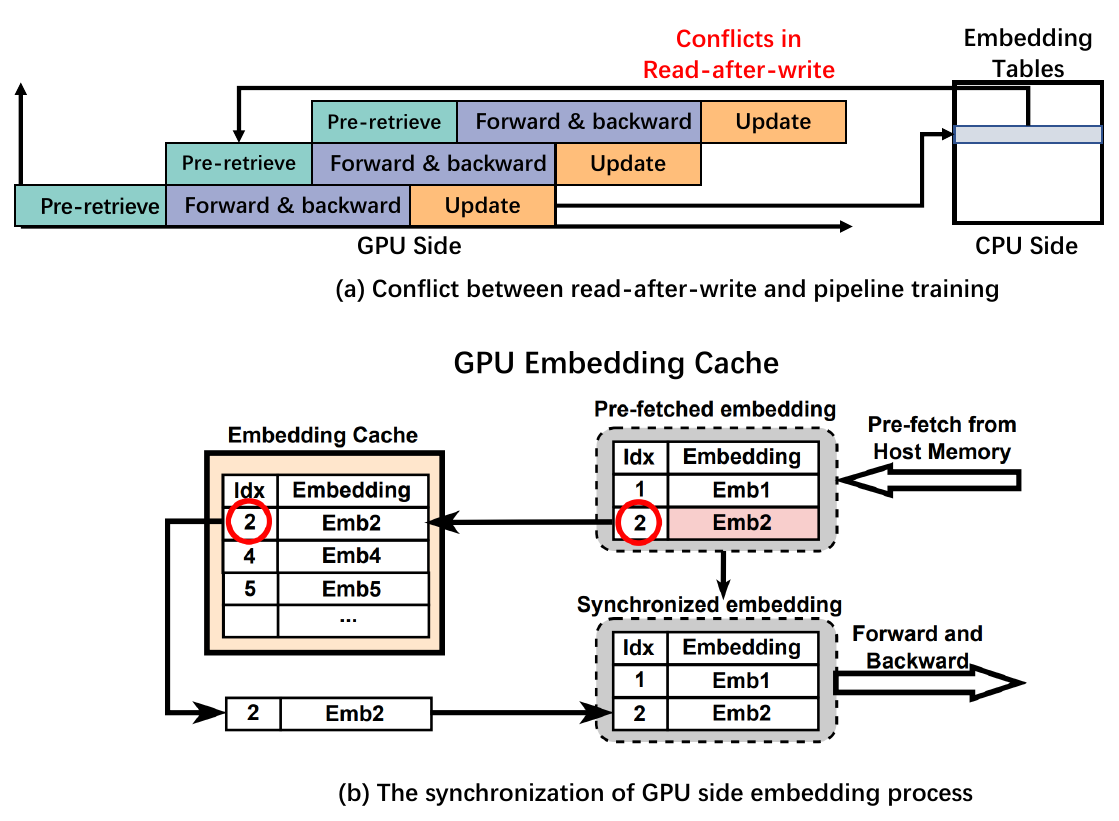}
    \vspace{-5pt}
    \caption{Using GPU-side embedded cache design to solve read-after-write conflicts: (a) The cause of read-after-write conflicts; (b) GPU-side embedded synchronization mechanism.}
    \label{fig: buffer}
    % \vspace{-5pt}
\end{figure}

\subsection{Read-after-write conflict optimization strategy}\label{sec: cache design}

Pipeline execution is a widely used technique to accelerate deep learning training on hierarchical memory architectures. In this design, embedding parameters are duplicated and stored on the CPU side, while forward and backward MLP computations are performed on the GPU.

However, such pipelining introduces Read After Write (RAW) conflicts. As illustrated in Fig.~\ref{fig: buffer}(a), while the MLP on GPU processes batch $i$, the embedding parameters for batch $i{+}1$ are pre-fetched. Since batch $i$ has not yet completed gradient updates, the embedding parameters used for batch $i{+}1$ may include \emph{stale values} that have not been refreshed by the latest updates.

To resolve this RAW hazard and ensure the correctness of embedding parameters for upcoming batches, we design a cache synchronization strategy with an adaptive filling policy. This mechanism dynamically updates GPU-side embeddings to ensure freshness and consistency.

As shown in Fig.~\ref{fig: buffer}(b), index and embedding configurations form a prefetch embedding batch. Upon receiving the prefetch stack, the GPU-side embedding cache performs index-wise lookups. If a hit occurs, the index is considered cached and can be reused. If the embedding has already been used in earlier rounds, it can be synchronized with the secondary cache (denoted as Emb2) to ensure consistency.

After embedding synchronization, all embeddings are guaranteed to be up to date and ready for training. We also introduce a lifecycle control mechanism to reduce the memory footprint of the integrated cache. This mechanism uses a cycle-based policy to control cache population.

When a training step completes, used embeddings are moved to the cache and marked for reassignment. The maximum length of the request queues, including the gradient queue and prefetch queue, is defined by the LC (Load Capacity) parameter. Both the gradient and prefetch stacks are dequeued from CPU memory according to their LC values. Once the LC value of an embedding reaches zero, it is evicted from the GPU cache.

This dynamic cache management strategy ensures efficient utilization of GPU memory and mitigates synchronization overhead in high-throughput pipeline training environments.

\section{Experiments and performance evaluation}
% We evaluate \Mname~on three real-world datasets and compare with several state-of-the-art DLRM frameworks. 

\subsection{Experimental setup}

\textbf{Implementation} 
We developed \textit{\Mname} based on \textit{DLRM}~\cite{naumov2019deep}, an open-source recommendation model training system suggested by Facebook, to demonstrate the advantages of our design. The PyTorch Wrapper facilitates the integration of the PyTorch framework with the C++/CUDA implementation of the Eff-TT table.  Moreover, the Eff-TT table can be seamlessly adapted to different PyTorch-based frameworks. Using it only requires replacing the PyTorch $\mathsf{nn.EmbeddingBag()}$ API.

\begin{table}[t]\small
\centering
\caption{Dataset Evaluation}
\vspace{-5pt}
\scalebox{0.8}{
\begin{tabular}{cccccc}
% \hline
\specialrule{.1em}{.05em}{.05em} 
\multirow{2}{*}{\textbf{Datasets}} & 
\multicolumn{2}{c}{\textbf{Input Features}} & 
\multicolumn{3}{c}{\textbf{Embedding Table}}  \\ \cline{2-6} 
&
\multicolumn{1}{c}{\textbf{Dense} } 
& \textbf{Sparse}  
& 
\multicolumn{1}{c}{\textbf{Rows}}  & 
\multicolumn{1}{c}{\textbf{Dimension}} & 
\multicolumn{1}{c}{\textbf{Size}}   \\ 
\hline

Avazu\cite{avazu-ctr-prediction}  & \multicolumn{1}{c}{1}  & 20  & \multicolumn{1}{c}{8.9M}  & \multicolumn{1}{c}{16}  & 0.55 GB \\ 

Criteo Terabyte\cite{WeTransfer}  & \multicolumn{1}{c}{13}  & 26  & \multicolumn{1}{c}{242.5M}  & \multicolumn{1}{c}{64}  & 59.2 GB   \\ 
Criteo Kaggle\cite{criteo-display-ad-challenge}  & \multicolumn{1}{c}{13}  & 26  & \multicolumn{1}{c}{30.8M} & \multicolumn{1}{c}{16}  & 1.9 GB  \\ 
IEEE118-Bus\cite{zimmerman2015matpower}& \multicolumn{1}{c}{6}  & 7  & \multicolumn{1}{c}{19.53M}  & \multicolumn{1}{c}{16}  & 1.22 GB   \\ 
% \hline
\specialrule{.1em}{.05em}{.05em} 
\end{tabular}
}
\label{table: Evaluation Dataset1}
\end{table}

\begin{algorithm}[t] \footnotesize
  \caption{Rec-AD trained with Eff-TT}
  \label{algo: train}
\SetAlgoLined
  \SetKwInOut{Input}{input}
  \Input{\textbf{Feature}: Perform maximum-minimum normalization on dense features and one-hot encoding on sparse features \\\textbf{label}: value to be regressed \\}
  
  \SetKwInOut{Output}{output}
  \Output{Expected target}
    %\tcc{Compute hot embedding threshold.

    Set the maximum number of iterations of the dataset: $epochs$, learning rate: $lr$;
    
    %\tcc{Iterate through all batches.}
    \For{$i$ $\mathbf{in}$ $range(epochs)$}{
        %\tcc{Global Information: get frequency based index.}
        For batch data: \\
            Sequential \\
            $dense features$ $\rightarrow$ $MLP$ mapping layer; \\
            $sparse features$ $\rightarrow$ $TT-embedding$ layer;\\
            %go through the TT-embedding layer, and then the two output features are fused, and finally the regression values are output through the MLP layer; \\
            $fused two outputs$; \\
            $outputs$ $=$ $fused$ $two$ $outputs$ $\rightarrow$ $MLP$ layer;\\
            %Calculate the $MSE$ loss(for PV datasets);\\
           Calculate $accuracy$ loss;\\
            %between the output and the real label(for PV datasets); \\
            Find the gradient $gradient$ of the model parameter loss; \\
            Update model parameters using gradient descent; \\
        %\tcc{Keep Hot embeddings' index.}
        %$\mathit{Fre\_batch}.\mathit{clamp}(min=\mathit{Hot\_thre})$ - $\mathit{Hot\_thre}$;
        
        %\tcc{Local Information: generate edges for $\mathit{Batch}$.}
        %$\mathit{Batch\_edges}$ = $\mathit{Fre\_batch}.\mathit{self\_combinations()}$;
        
        %\tcc{Append to $\mathit{Edge\_list}[\ ]$.}
        %$\mathit{Edge\_list}.\mathit{append}(\mathit{Batch\_edges})$;
    }
    %The output layer uses Sigmoid when the training dataset is a recommender system dataset and MSE loss function when the dataset are PV datasets.
\end{algorithm}

%此处添加光伏数据集验证

% \begin{figure} [t] \small
%     \centering
%     \includegraphics[width=0.7\linewidth]{figures/sitemap.pdf}
%     \vspace{-5pt}
%     \caption{PV sites resource allocation.}
%     \label{fig: sitemap}
%     \vspace{5pt}
% \end{figure}

\subsection{Datasets and Evaluation Settings}
\label{sec: datasets}

We evaluate Rec-AD on four commonly used datasets:

\textbf{IEEE 118-Bus}~\cite{zimmerman2015matpower} is a widely adopted power system benchmark dataset. According to the configuration described in reference~\cite{LI2021119505}, each sample consists of 7 sparse features (e.g., categorical, topological) and 6 dense features (e.g., voltage magnitude, power). Details can be found in reference~\cite{LI2021119505}. While they only considers features such as line power flow, bus injection, voltage magnitude, and phase angle, this paper additionally includes unstructured features such as node topology and load characteristics.

The IEEE 118-bus dataset contains a total of 24,800 samples, with 20,000 labeled as normal and 4,800 labeled as FDIA (False Data Injection Attack) samples. After data preprocessing and feature selection, the dataset is split into training and testing sets.

\textbf{Avazu}~\cite{avazu-ctr-prediction} is a dataset collected over 11 days from Avazu ad logs for click-through rate (CTR) prediction. Each sample includes 20 categorical (sparse) features and 1 numerical (dense) feature. It is a standard benchmark for evaluating DLRM-style recommendation models.

\textbf{Criteo Terabyte}~\cite{WeTransfer} is the largest publicly available benchmark dataset for DLRM, containing over 4 billion samples spanning 24 days. Each record is composed of 13 continuous (dense) features and 26 categorical (sparse) features, totaling 39 fields. The embedding table size reaches approximately 59.2 GB, which exceeds the memory capacity of most commercial GPUs.

\textbf{Criteo Kaggle}~\cite{criteo-display-ad-challenge} is a subset of the Criteo Terabyte dataset used in the Criteo Display Advertising Challenge on Kaggle. It follows the same feature format and contains a representative 7-day sample of Criteo ad traffic.

Detailed dataset statistics are summarized in Table~\ref{table: Evaluation Dataset1}. Notably, the size of the embedding table in Criteo Terabyte already poses a challenge to memory-constrained environments. In industrial applications, datasets are often even larger and more complex~\cite{zhao2020distributed}, which further motivates the need for a scalable and memory-efficient solution like Rec-AD.

\subsection{Baseline Systems for Comparison}
\label{sec: baselines}

In addition to the vanilla DLRM framework, we select four state-of-the-art DLRM-based systems that incorporate optimizations at either the algorithmic or system level. These baselines serve as strong points of comparison to evaluate the effectiveness of Rec-AD.

\begin{enumerate}
    \item \textbf{TT-Rec}~\cite{yin2021tt}, proposed by Facebook, performs algorithm-level optimization by applying Tensor Train (TT) decomposition to compress large embedding tables. In our evaluation, we integrate the TT-Rec embedding compression API directly into the DLRM framework to ensure fair comparison. During end-to-end training, both Rec-AD and TT-Rec compress embedding tables with over one million rows, while smaller embedding tables are left uncompressed.

    \item \textbf{FAE (Fast and Efficient)}~\cite{ebrahimzadeh2021accelerating} focuses on system-level design. It utilizes host memory to manage large embedding tables and offloads frequently accessed embeddings to the GPU. By minimizing CPU–GPU communication overhead, FAE allows most training batches to be processed entirely on the GPU.

    \item \textbf{HugeCTR}~\cite{hugectr}, developed by NVIDIA, is a production-grade, high-performance recommendation training system. It adopts data-parallel MLP training and model-parallel embedding table processing. HugeCTR supports distributed training across multiple GPUs and nodes. For large-scale embeddings, it scales up the number of GPUs to fit the entire embedding table into GPU HBM.

    \item \textbf{TorchRec}~\cite{TorchRec}, a recent release from Facebook, implements large-scale embedding operations within the DLRM framework. It introduces 4D parallelism~\cite{mudigere2021softwarehardware}, which combines table-wise, row-wise, column-wise, and data parallelism. For model-parallel training, TorchRec partitions large embedding tables via row or column sharding across multiple devices.
\end{enumerate}

These baseline systems reflect diverse optimization strategies—TT-Rec emphasizes low-rank compression, FAE leverages hierarchical memory systems, HugeCTR excels in scalable GPU computation, and TorchRec offers fine-grained model parallelism. Comparing Rec-AD with these systems enables us to analyze its effectiveness across multiple dimensions including memory efficiency, computational speed, and scalability.

\subsection{Experimental Platform and Implementation}
\label{sec: experiment-platform}

Our primary evaluation platform is an AWS p3.8xlarge instance, equipped with 4 NVIDIA Tesla V100 GPUs, 239 GB of CPU memory, and a 2.30 GHz Intel Xeon processor. For additional comparison, we also evaluate overall performance on an AWS g4dn.12xlarge instance, which is configured with 4 NVIDIA Tesla T4 GPUs, 192 GB of CPU memory, and a 2.50 GHz Intel CPU.

We adopt NVIDIA NVTabular~\cite{NvTabular} as our data preprocessing pipeline. NVTabular also provides a high-performance data loader for streaming DLRM training data from disk to memory. To ensure a fair comparison across all baselines, we replace the NVTabular data loader in the benchmark frameworks with the default PyTorch DataLoader for all training experiments.

The implementation of the Eff-TT table in Rec-AD utilizes the cuBLAS library~\cite{cublas} for performing high-throughput batched General Matrix Multiply (GEMM) operations. This ensures that the performance-critical TT embedding lookup and gradient update procedures are executed with GPU-optimized matrix primitives.

\subsection{Performance Analysis}

\begin{table}[htbp]\small
  \centering
  \begingroup
  \caption{Comparison of FDIA detection training time and detection performance on IEEE118-Bus}
  \scalebox{0.66}{
  \begin{tabular}{lcccccc}
    \toprule
   \textbf{Models} &  & \textbf{Training} &  & & \textbf{Detection } &  \\ 
   &&\textbf{time}&&&\textbf{performance}\\
   \cline{2-4} \cline{5-7}
   %\hline
     & \textbf{CPU} & \textbf{1GPU} & \textbf{4 GPU} & \textbf{Accuracy (\%)} & \textbf{Recall (\%)} & \textbf{F1-Score (\%)} \\
    \midrule
    DLRM(baseline)  & 1.00 & 1.00 & 1.00 & 94.1   & 92.2  & 92.1 \\
    TT-Rec      & 0.90 & 0.82 & 0.68 & 96.8      & 95.3      & 95.8  \\
    Rec-AD      & 0.82 & 0.74 & 0.62 & 97.5       & 96.2      & 96.3  \\
   % MEC  & 1.00 & 1.08 & 1.17 & 98.2       & 97.1      & 96.5  \\
    \bottomrule
  \end{tabular}%
  \label{table:time&acc}
  }
  \endgroup
\end{table}

\subsection{Evaluation Metrics and Detection Performance}
\label{sec: evaluation-metrics}

In the FDIA detection task, three primary metrics are used to evaluate model performance: Accuracy, Recall, and F1-Score. 

\textit{Accuracy} measures the overall correctness of the model's predictions and is defined as the proportion of correctly classified samples among all samples. While accuracy provides a general sense of classification capability, it can be misleading in imbalanced datasets, such as FDIA detection scenarios where attack samples are significantly fewer than normal ones. In such cases, accuracy may fail to reflect the model's true effectiveness in identifying attacks.

In contrast, \textit{Recall} emphasizes the ability to detect positive (attack) samples and is calculated as the proportion of true attacks that are correctly identified. High recall indicates that the model is effective in reducing \textit{false negatives}, which is crucial for FDIA detection, as undetected attacks may cause cascading failures or safety risks in the power grid. However, a high recall may lead to an increase in \textit{false positives}, which could compromise the model's stability and precision.

To strike a balance between precision and recall, the F1-Score is widely adopted. It is the harmonic mean of precision and recall, offering a comprehensive metric that reflects the model's ability to reduce both false positives and false negatives. In FDIA detection, a high F1-Score implies that the model can accurately detect attacks while maintaining a low false alarm rate, thus ensuring robust protection in complex power system environments.

\subsection{Comparative Results and Analysis}
\label{sec: results}

During model training, we compare the baseline DLRM framework against TT-Rec and our proposed Rec-AD, both of which apply TT decomposition for embedding compression. To facilitate an intuitive comparison of training efficiency, we normalize the training time of the baseline DLRM model to 1.0. Detailed results are presented in Table~\ref{table:time&acc}.

Experimental results show that incorporating TT decomposition not only reduces computational complexity but also improves detection accuracy and model robustness in FDIA tasks. These gains indicate that the compressed TT-based embedding layer better captures essential structural and semantic properties of input features, thereby enhancing overall detection performance.

In practice, the training time of an FDIA detection model is influenced by multiple factors, including network architecture, parameter size, dataset scale, and hardware acceleration. For the DLRM architecture, training primarily involves large-scale matrix multiplication and vector operations, which are highly parallelizable on GPUs. As a result, GPU acceleration typically yields substantial speedups. However, in the IEEE 118-Bus environment, where the dataset is relatively small, the acceleration effect may be less pronounced.

The DLRM+TT approach integrates both dense and sparse features while compressing the embedding layer using TT decomposition. This reduces model complexity and leads to a 1.5$\sim$2$\times$ training speedup compared to the baseline on GPU platforms. Moreover, due to its parameter efficiency and improved feature modeling, this method mitigates overfitting and enhances the extraction of critical patterns.

Rec-AD performs competitively in FDIA detection, achieving approximately 97.5\% Accuracy, 96.2\% Recall, and an F1-Score of 96.3\%. These results highlight the effectiveness of Rec-AD in balancing precision and recall, while maintaining high overall classification accuracy.

% In summary, Rec-AD significantly outperforms the standard DLRM framework in terms of both training efficiency and detection performance. Compared to other strong baselines such as TT-Rec and MEC, Rec-AD offers competitive or superior results, making it a practical and robust solution for real-time security monitoring in smart grid environments.

\subsection{Computational efficiency analysis} 
Here we present the training throughput of Rec-AD and evaluate it against other baseline models. To demonstrate the advantages of Rec-AD in various scenarios, we create two main evaluation environments:
1) DLRM training on a single GPU or with limited GPU resources;
2) DLRM training on multiple GPUs.

\begin{table}[t] \small
\centering
% \vspace{-5pt}
\caption{Table footprint embedding comparison}
\vspace{-5pt}
\scalebox{1}{
\begin{tabular}{l r r c}
\specialrule{.1em}{.05em}{.05em} 
% \textbf{Dataset} & \textbf{EMB table} & \textbf{TT table} & \textbf{CR} 
\textbf{DataSets} & \textbf{DLRM} & \textbf{\Mname} & \multicolumn{1}{c}{\textbf{Compression ratio}}
\\ \hline 

{Avazu} & 0.55GB & 87.6MB & ~~6.22$\times$ \\ 
% \hline
{Terabyte}  & 59.2GB & 797.9MB & 74.19$\times$  \\
% \hline
{Kaggle}  & 1.9GB & 258.2MB & ~~7.29$\times$  \\ 

{IEEE118-Bus}  & 1.22GB & 235.7MB & ~~5.33$\times$  \\
% {\Mname~}  & 83.52& 81.96 & 78.52 \\ 
\specialrule{.1em}{.05em}{.05em} 
\end{tabular}
}
\label{table: compression ratio}
% \vspace{5pt}
\end{table}

% \subsection{End-to-End Training Time and Multi-GPU Scalability}
% \label{sec:end2end-performance}

\begin{figure*} [t] \small
    \centering
    \includegraphics[width=0.9\linewidth]{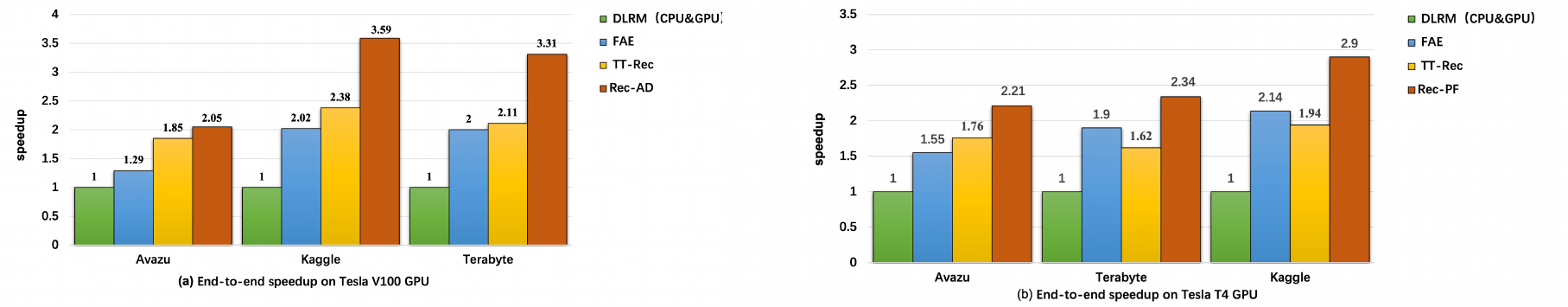}
    % \vspace{-5pt}
    \caption{With limited GPU resources, single GPU can achieve end-to-end training speed-up ($\times$).}
    \label{fig: end2end}
    \vspace{-5pt}
\end{figure*}

To simulate a resource-constrained environment, we perform experiments using a single GPU on each platform. In this setting, both DLRM and FAE store embedding table parameters in host memory, while TT-Rec and Rec-AD store compressed embeddings directly in GPU HBM.
For both Tesla V100 and Tesla T4 GPUs, we adopt a batch size of 4096. The TT embedding dimension is set to 128 for V100 and 64 for T4. Fig.~\ref{fig: end2end} presents the end-to-end training time comparison between Rec-AD and all baseline models. 
Across all systems and datasets, Rec-AD consistently achieves the best overall performance. On the Tesla V100, Rec-AD achieves an average speedup of 3$\times$ over the baseline DLRM. Since all parameters can be held within a single GPU and the Eff-TT embedding table occupies very little memory, there is minimal communication overhead between CPU and GPU, resulting in substantial gains.

Compared to FAE, Rec-AD is on average 1.5$\times$ faster. FAE stores frequently accessed embeddings on the GPU, but approximately 25\% of batches contain cold embeddings that still require CPU-side access and computation, limiting FAE’s performance ceiling.
Additionally, Rec-AD outperforms TT-Rec by an average of 1.4$\times$. This improvement stems from Rec-AD’s series of low-level optimizations that reduce the computational cost of TT embedding lookup and backpropagation, leading to better throughput during training.

% \subsection{Multi-GPU Training Throughput}
% \label{sec:multi-gpu}

\begin{figure} [t] \small
    \centering
    \includegraphics[width=0.90\linewidth]{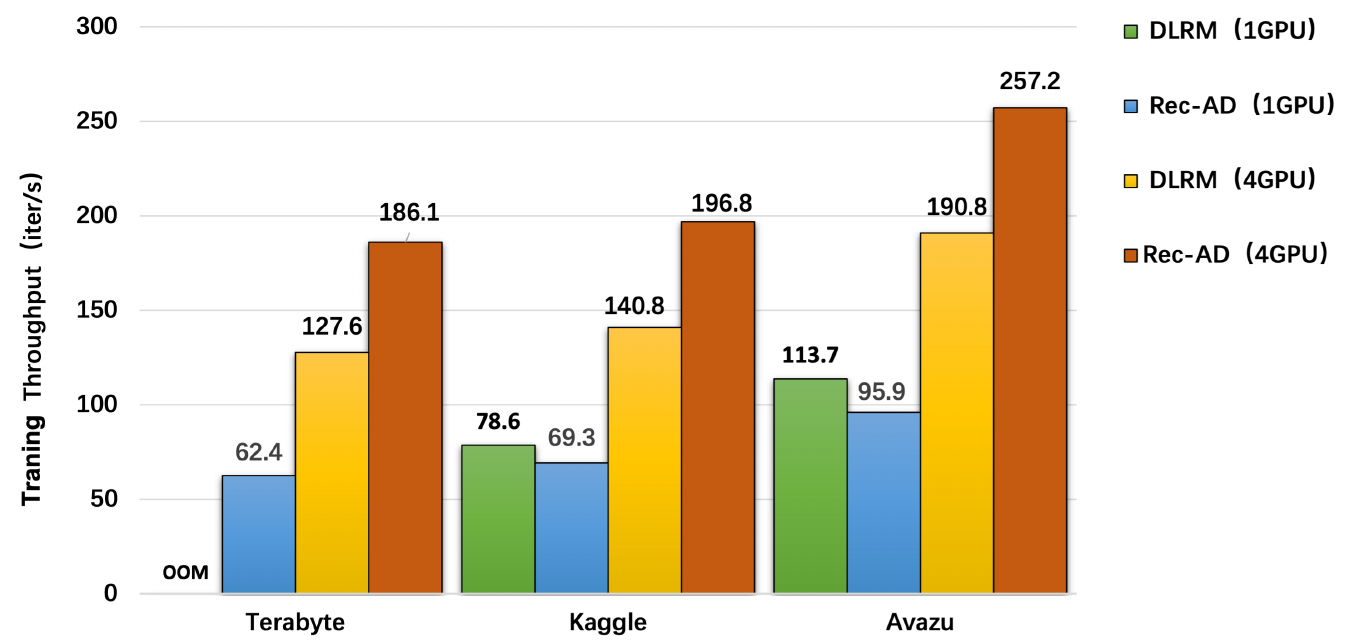}
    \vspace{-5pt}
    \caption{Training throughput with multiple GPUs enabled}
    \label{fig: multi GPU}
    \vspace{5pt}
\end{figure}

To explore scalability, we compare training throughput on the AWS p3.8xlarge platform using 1 GPU vs. 4 GPUs for both Rec-AD and DLRM. The results are shown in Fig.~\ref{fig: multi GPU}.
Rec-AD (4 GPUs) achieves 1.4$\times$ higher training throughput than DLRM (4 GPUs). This is mainly due to the compact memory footprint of the Eff-TT embedding table, which enables fully data-parallel embedding training across all GPUs with efficient synchronization and parameter replication.

In contrast, DLRM distributes embedding tables across GPUs in a model-parallel fashion, which necessitates frequent peer-to-peer communication between devices. This overhead limits scalability and efficiency as the number of GPUs increases.
On the 1 GPU setting, DLRM slightly outperforms Rec-AD in raw throughput. This is expected because while Rec-AD's tensorized embedding significantly reduces memory usage, it introduces additional computational overhead (as shown in Table~\ref{table: compression ratio}). Nevertheless, Rec-AD demonstrates superior scalability in large-scale DLRM training, offering a promising solution for industrial-scale deployment scenarios.

\begin{figure} [t] \small
    \centering
    \includegraphics[width=0.90\linewidth]{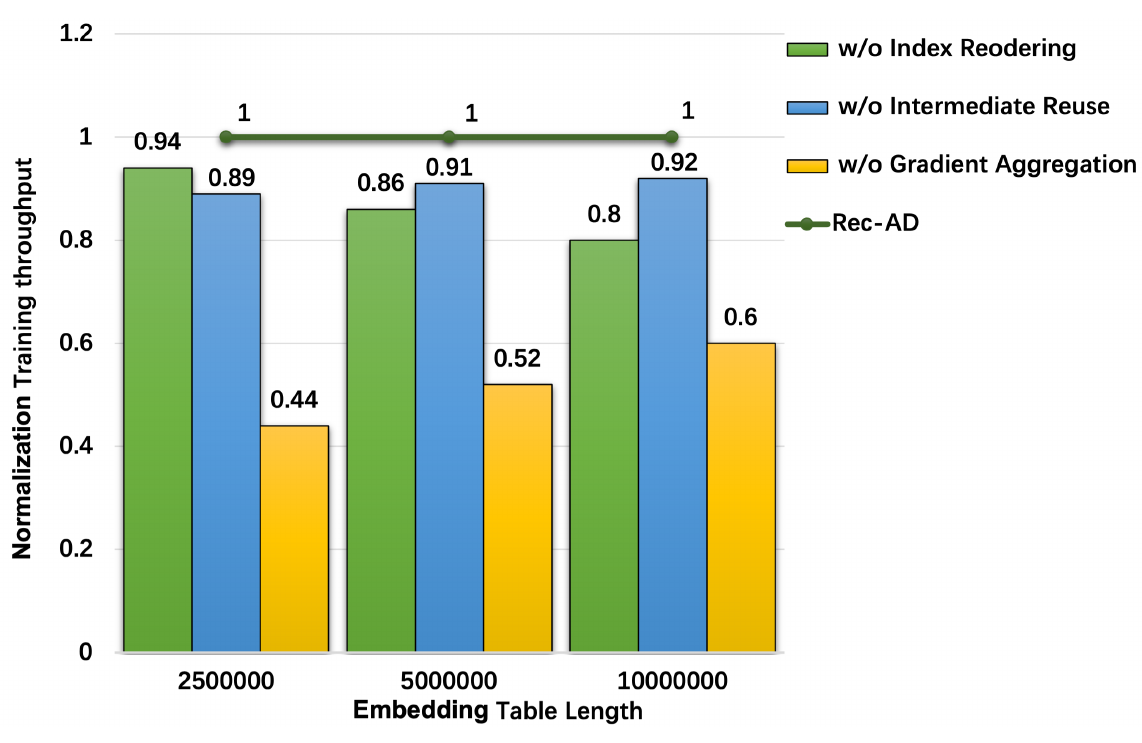}
    \vspace{-5pt}
    \caption{Eff-TT table optimization decomposition}
    \label{fig: single_table_breakdown}
    \vspace{5pt}
\end{figure}

\subsection{Training with Large Embedding Tables}
\label{sec:large-table}

To demonstrate the advantage of Rec-AD in handling ultra-large embedding tables, we constructed an embedding table with 40 million rows and 128 embedding dimensions, occupying approximately 19 GB—exceeding the 16 GB HBM capacity of our GPUs.

\begin{figure} [t] \small
    \centering
    \scalebox{0.35}{\includegraphics{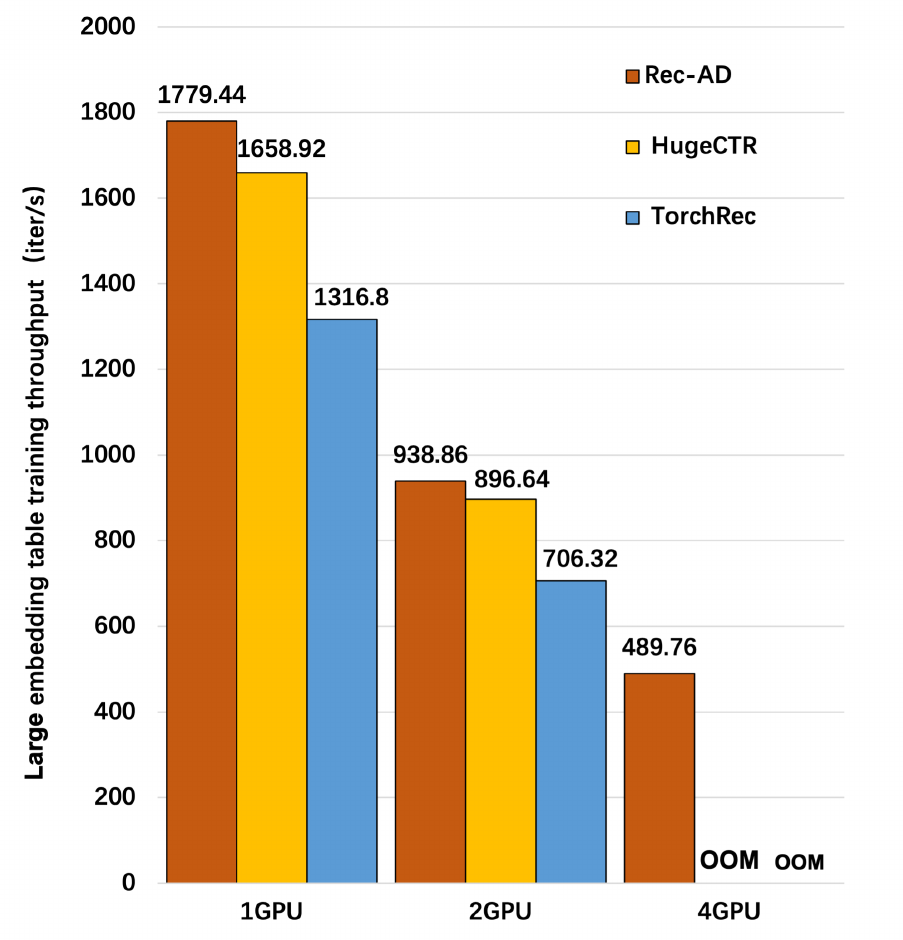}}
    \vspace{-5pt}
    \caption{Throughput comparison of HugeCTR and TorchRec for training a single large embedding table}
    \label{fig: single_large_table}
    \vspace{-5pt}
\end{figure}

We benchmarked three models on this large table setup: Rec-AD, HugeCTR, and TorchRec, each trained with varying numbers of GPUs. As shown in Fig.~\ref{fig: single_large_table}, Rec-AD achieved 1.35$\times$ speedup over TorchRec and 1.07$\times$ over HugeCTR.

Due to the limited HBM capacity, HugeCTR distributes embedding table parameters across multiple GPUs, while TorchRec uses column-wise sharding to partition the table and execute model-parallel training. However, these approaches incur significant inter-GPU communication overhead, particularly during forward-pass embedding synchronization and backward-pass gradient aggregation.

In contrast, Rec-AD minimizes memory footprint via the Eff-TT table, enabling the training of ultra-large embeddings within a single GPU. During the forward pass, Rec-AD adopts data-parallel training on the compressed embedding representation to avoid synchronization overhead and improve system throughput.

%\ZY{Figure 10-13 can be compressed to shorten pages}

\subsection{Accuracy and Loss Comparison}
\label{sec:accuracy-eval}

We evaluated Rec-AD on multiple datasets to assess whether tensorization impacts model accuracy. All models were trained for 100K iterations on the Criteo Terabyte dataset and 5 epochs on the Criteo Kaggle and Avazu datasets.

As summarized in Table~\ref{table: Accuracy}, Rec-AD achieves comparable accuracy to baseline models across all three datasets. The slight drop in accuracy for Rec-AD on the Terabyte dataset ($<$0.1\%) is due to its aggressive compression via the Eff-TT table, which significantly reduces parameter space.

However, this minor loss in accuracy is negligible and acceptable in real-world applications, particularly when considering the substantial improvement in memory efficiency and training throughput. These results confirm that Rec-AD maintains competitive model quality while enabling scalable training on ultra-large datasets.

% \begin{figure*} [t] \small
%     \centering
%     \includegraphics[width=0.95\linewidth]{figures/PV_predicted1.pdf}
%     % \vspace{-5pt}
%     \caption{UNISOLAR太阳能发电数据集中42个站点的预测值与实际值的4个站点}
%     \label{fig: PV_predicted}
%     \vspace{5pt}
% \end{figure*}

\begin{table}[t] \small
\centering
% \vspace{-5pt}
\caption{Comparison of prediction accuracy}
\vspace{-7pt}
\scalebox{0.9}{
\begin{tabular}{l|c c c}
\specialrule{.1em}{.05em}{.05em} 
\diagbox{\textbf{Models}}
{\textbf{Datasets}} &  \textbf{Avazu} & \textbf{Criteo Terabyte} & \textbf{Criteo Kaggle} 
\\ \hline 
\\[-0.9em]
{DLRM} & 83.53 & 81.96 & 78.53 \\ 
{TT-Rec}  & 83.51 & 81.86 & 78.51 \\
{FAE}  & 83.53 & 81.94 & 78.52 \\ 
\hline
\\[-0.9em]
{\textbf{\Mname~}}  & \textbf{83.51} & \textbf{81.90} & \textbf{78.50} \\ 
% {\Mname~(CPU-GPU)}  & 83.52& 81.96 & 78.52 \\ 
% {\Mname~}  & 83.52& 81.96 & 78.52 \\ 
\specialrule{.1em}{.05em}{.05em} 
\end{tabular}
}
\label{table: Accuracy}
% \vspace{5pt}
\end{table}

\begin{figure} [h] \small
    \centering
    \includegraphics[width=0.8\linewidth]{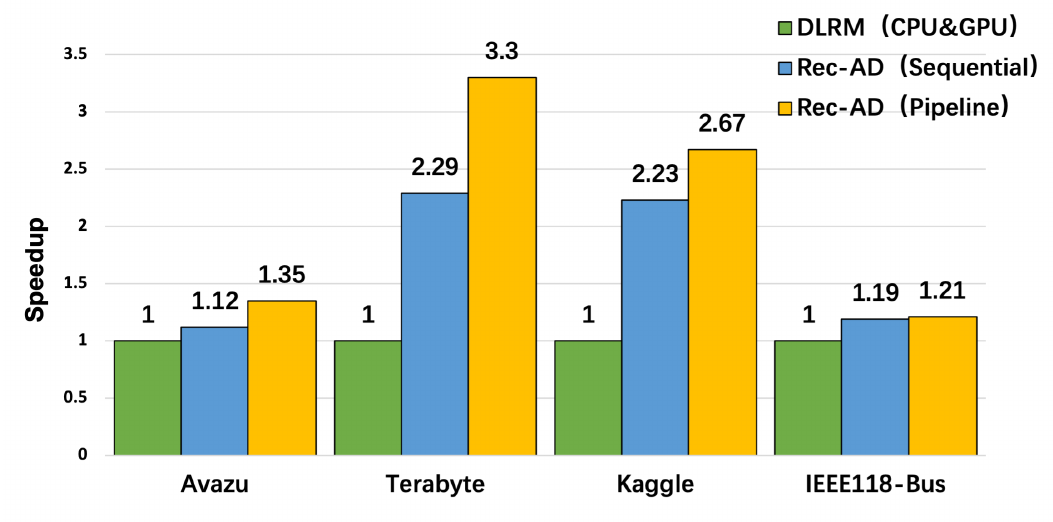}
    \vspace{-5pt}
    \caption{Speedup of partial optimization training vs. overall optimization training ($\times$)}
    \label{fig: CPUGPU}
    % \vspace{-5pt}
\end{figure}

%\vspace{-3pt}
\subsection{Ablation Study: Impact of Individual Optimizations}
\label{sec:ablation}

To evaluate the individual contributions of each optimization in Rec-AD, we trained three embedding tables of varying sizes, ranging from 2.5 million to 10 million rows. We then selectively disabled one optimization at a time and measured the resulting training throughput. The results, shown in Fig.~\ref{fig: single_table_breakdown}, highlight that forward-pass gradient aggregation has the most significant impact on performance.

When gradient aggregation is disabled, the training throughput drops by approximately 52\%. This is due to the forward TT table involving significantly less computation compared to the backward TT table, making forward aggregation a key performance contributor. Additionally, index reordering and intermediate result reuse also show meaningful improvements. Disabling these two optimizations causes a performance degradation of 13\% and 10\% on average, respectively.

As the size of the embedding table increases, the benefit of index reordering becomes more pronounced. Larger tables lead to more scattered index accesses, offering greater potential for reordering to improve locality.

\subsection{Pipeline Training System}
\label{sec:pipeline}

To demonstrate the advantage of our TT-based pipeline training system, we compress the largest embedding table into an Eff-TT table stored in GPU HBM, while placing the remaining embeddings in host memory. To isolate the effect of pipelining, we set the prefetch queue length to 1, effectively disabling the pipeline, and compare the performance to sequential training and other baselines.

As shown in Fig.~\ref{fig: CPUGPU}, Rec-AD (Pipeline) achieves an average 2.44$\times$ speedup over DLRM, showcasing the benefits of overlapping parameter updates and embedding lookups on the GPU with MLP training on the CPU. The use of prefetch and gradient queues significantly reduces CPU-GPU communication overhead.

Moreover, Rec-AD (Pipeline) outperforms Rec-AD (Sequential) by an average of 1.30$\times$. When the prefetch queue length is set to 1 in Rec-AD (Sequential), the pipeline degrades into a purely sequential execution. In this case, GPU workers must wait for the CPU to complete parameter updates before accessing the next batch—emphasizing the necessity of pipelined design and embedding cache mechanisms.

The pipeline training architecture enhances computational efficiency, enabling more resilient FDIA detection across diverse smart grid scenarios. By reducing reliance on centralized servers, this system lays the foundation for distributed security monitoring and promotes scalable, edge-friendly deployment in future smart grid infrastructures.

\subsection{Engineering application analysis}
\begin{table}[htbp]\small
\centering
\caption{Performance comparison between DLRM and Rec-AD on a 100MB dataset (Batch Size = 1)}
\scalebox{0.6}{
\begin{tabular}{lccc}
\toprule
 & \textbf{DLRM } & \textbf{Rec-AD } & \textbf{Improvement} \\
 &@100MB&@100MB&\textbf{instructions}\\
\midrule
Batch Size           & 1               & 1               & Streaming real-time detection  \\
&&& configuration(industrial typical)\\
&&&\\
Deployment platform            & RTX 2060  & RTX 2060          & Industrial-grade  \\
&&&GPU supports FP32/FP16\\
&&&\\
Single Detection Latency       & 25 ms           & 21.5 ms         & ↓14\% \\
&&&More real-time response\\
&&&\\
Throughput(TPS)                 & 40 p/s         & 46.5 p/s       & ↑16\% \\
&&&Higher concurrent processing capabilities\\
&&&\\
100MB Total Time  & 3.47 hours        & 3.0 hours       & ↓0.47 hours \\
&&&Faster batch testing\\
&&&\\
Peak memory usage (GPU)           & 320 MB          & 210 MB          & ↓34\% \\
&&&Adapting to edge device operation\\
&&&\\
Model deployment volume                 & 180 MB          & 95 MB           & ↓47\% \\
&&&Facilitates embedded OTA deployment\\
&&&\\
Industrial deployment compatibility               & Medium             & Strong               & Support embedded \\
&&&energy-saving deployment\\
\bottomrule
\end{tabular}
\label{tab:dlrm_rec_pf_batch1_comparison}
}
\end{table}

In industrial-scale smart grid environments, real-time responsiveness and precise defense against false data injection attacks (FDIAs) are critical. Particularly under large-scale data conditions, inference latency, resource consumption, and system throughput become decisive factors for effective model deployment. Based on the IEEE 118-bus system, we constructed a 100MB-scale FDIA detection environment to compare the engineering performance of the original DLRM model and our proposed Rec-AD optimization under a Batch Size = 1 (streaming inference) scenario.

As shown in Table~\ref{tab:dlrm_rec_pf_batch1_comparison}, the Rec-AD model outperforms the baseline DLRM model across several key metrics. Under identical batch size settings, Rec-AD leverages Tensor Train-based embedding compression and pipeline architecture to reduce inference latency from 25\,ms (DLRM) to 21.5\,ms, achieving a 14\% latency reduction, thereby significantly improving responsiveness and real-time capability. Correspondingly, the throughput (TPS) increases from 40 samples/s to 46.5 samples/s—an improvement of 16\%, better supporting high-frequency sampling scenarios.

In terms of resource consumption, Rec-AD also demonstrates superior efficiency. Thanks to its parameter compression mechanism, runtime memory usage is reduced from 320\,MB to approximately 210\,MB, alleviating the resource load by over 34\%. Moreover, the model's deployment size shrinks from 180\,MB to 95\,MB—almost halved, making it especially suitable for deployment on embedded edge terminals or resource-constrained industrial nodes such as substation intelligent terminals and edge servers.

Further analysis reveals that when continuously processing 100MB-scale real-time data, the DLRM model requires approximately 3.47 hours to complete the detection task, whereas Rec-AD shortens the total processing time to 3.0 hours, saving nearly 0.5 hours in detection cycles. This efficiency gain is particularly valuable in daily batch processing and rolling prediction tasks.

In summary, Rec-AD significantly enhances system responsiveness, resource efficiency, and edge compatibility while maintaining detection accuracy. It is particularly well-suited for deployment in industrial smart grid environments aimed at security monitoring, fast fault identification, and dynamic situational awareness, demonstrating strong practical engineering value.

\section{Conclusion}
We proposed \Mname, a cost-efficient framework for detecting False Data Injection Attacks (FDIAs) in smart grids under limited GPU resources. \Mname~integrates Tensor Train (TT)-based embedding compression with deep recommendation models, significantly reducing memory footprint and computational overhead while maintaining detection accuracy.

By leveraging both global and local index patterns, \Mname~employs an index-aware reordering strategy to improve TT embedding computation efficiency. The TT decomposition compresses high-dimensional embeddings into low-rank cores, accelerating both training and inference. These enhancements enable faster model updates and real-time response—crucial for the stability and adaptability of cyber-physical energy systems.
In addition, a TT-based pipeline training mechanism reduces CPU–GPU data transfer bottlenecks, enhancing throughput and enabling deployment on resource-constrained platforms. Experiments demonstrate that Rec-AD outperforms state-of-the-art DLRM training systems in runtime and scalability, even with limited high-bandwidth memory.

% \textcolor{blue}{Looking ahead, we plan to explore hybrid probabilistic models, adaptive data preprocessing, and ensemble learning to further boost scalability and generalization. In the next chapter, we will extend Rec-AD to a federated learning setting, enabling secure and efficient FDIA detection across distributed energy systems.}

% \section*{Acknowledgments}
% This should be a simple paragraph before the References to thank those individuals and institutions who have supported your work on this article.

\bibliography{Test-bibtex}

\bibliographystyle{IEEEtran}

\vfill

\end{document}